\documentclass{article}
\usepackage{iclr2026_conference, times}

\usepackage{amsmath,amsfonts,bm}

\def\eqref#1{equation~\ref{#1}}

\def\1{\bm{1}}

\DeclareMathAlphabet{\mathsfit}{\encodingdefault}{\sfdefault}{m}{sl}
\SetMathAlphabet{\mathsfit}{bold}{\encodingdefault}{\sfdefault}{bx}{n}

\DeclareMathOperator*{\argmax}{arg\,max}

\usepackage{hyperref}
\usepackage{url}
\usepackage{graphicx}
\usepackage{algorithm}
\usepackage{algorithmic}
\usepackage[most]{tcolorbox}
\usepackage{booktabs}
\usepackage{subcaption}
\usepackage{wrapfig}
\usepackage{enumitem}
\newcount\Comments  %
\definecolor{darkgreen}{rgb}{0,0.5,0}
\definecolor{darkred}{rgb}{0.7,0,0}
\definecolor{teal}{rgb}{0.1,0.6,0.7}
\definecolor{blue}{rgb}{0.0,0.1,0.9}
\definecolor{orange}{rgb}{1.,0.7,0.0}
\definecolor{palegreen}{rgb}{0.7,0.7,0.0}
\definecolor{lightblue}{rgb}{0.70, 0.80, 0.89}
\definecolor{violet}{rgb}{0.50, 0.16, 0.88}
\definecolor{babyblue}{rgb}{0.00, 0.88, 0.88}
\definecolor{electricpurple}{rgb}{0.75, 0.0, 1.0}

\newcommand{\kibitz}[2]{\ifnum\Comments=1{{\textcolor{#1}{\textsf{\footnotesize [#2]}}}}\fi}

\newcommand{\method}{\textsc{MiGrATe}}  %

\newcommand{\task}{\ensuremath{\mathcal{T}}}
\newcommand{\taskprompt}{\ensuremath{P_{\task}}}
\newcommand{\group}{\ensuremath{\mathcal{G}}}
\newcommand{\database}{\ensuremath{\mathcal{D}}}
\newcommand{\budget}{\ensuremath{B}}

\newcommand{\groupsize}{\ensuremath{N}}
\newcommand{\nonline}{\ensuremath{\alpha}}
\newcommand{\ngreedy}{\ensuremath{\beta}}
\newcommand{\nnbr}{\ensuremath{\gamma}}
\newcommand{\nsprompt}{\ensuremath{P_\text{NS}}}
\newcommand{\obsonline}{\ensuremath{\mathcal{O}_\text{online}}}
\newcommand{\obsgreedy}{\ensuremath{\mathcal{O}_\text{greedy}}}
\newcommand{\obsns}{\ensuremath{\mathcal{O}_\text{NS}}}
\newcommand{\bestresult}[1]{\boldmath{$#1$}}
\newcommand{\secondbestresult}[1]{\underline{$#1$}}

\title{\method{}: Mixed-Policy GRPO for Adaptation at Test-Time}

\author{Peter Phan\thanks{These authors contributed equally.}\,, Dhruv Agarwal\footnotemark[1]\,, Andrew McCallum\quad\quad\quad\quad\quad\quad\\
University of Massachusetts Amherst \\
Amherst, MA 01003, USA \\
\texttt{\{pkphan, dagarwal, mccallum\}@cs.umass.edu} \\
\And
Kavitha Srinivas, Horst Samulowitz, Pavan Kapanipathi \\
IBM Research\\
\texttt{\{kavitha.srinivas, samulowitz, kapanipa\}@ibm.com} \\
}

\arxivtrue %
\begin{document}

\maketitle

\begin{abstract}
Large language models (LLMs) are increasingly being applied to black-box optimization tasks, from program synthesis to molecule design. Prior work typically leverages in-context learning to iteratively guide the model towards better solutions. Such methods, however, often struggle to balance exploration of new solution spaces with exploitation of high-reward ones. Recently, test-time training (TTT) with synthetic data has shown promise in improving solution quality. However, the need for hand-crafted training data tailored to each task limits feasibility and scalability across domains.
To address this problem, we introduce \method{}---a method for \emph{online} TTT that uses GRPO as a \emph{search} algorithm to adapt LLMs at inference without requiring external training data. \method{} operates via a mixed-policy group construction procedure that combines on-policy sampling with two off-policy data selection techniques: greedy sampling, which selects top-performing past completions, and neighborhood sampling (NS), which generates completions structurally similar to high-reward ones. Together, these components bias the policy gradient towards exploitation of promising regions in solution space, while preserving exploration through on-policy sampling.
We evaluate \method{} on three challenging domains---word search, molecule optimization, and hypothesis+program induction on the Abstraction and Reasoning Corpus (ARC)---and find that it consistently outperforms both inference-only and TTT baselines, demonstrating the potential of online TTT as a solution for complex search tasks without external supervision.

\end{abstract}

\section{Introduction}
\label{sec:intro}

\begin{figure}[ht]
    \centering
    \includegraphics[height=0.33\textwidth]{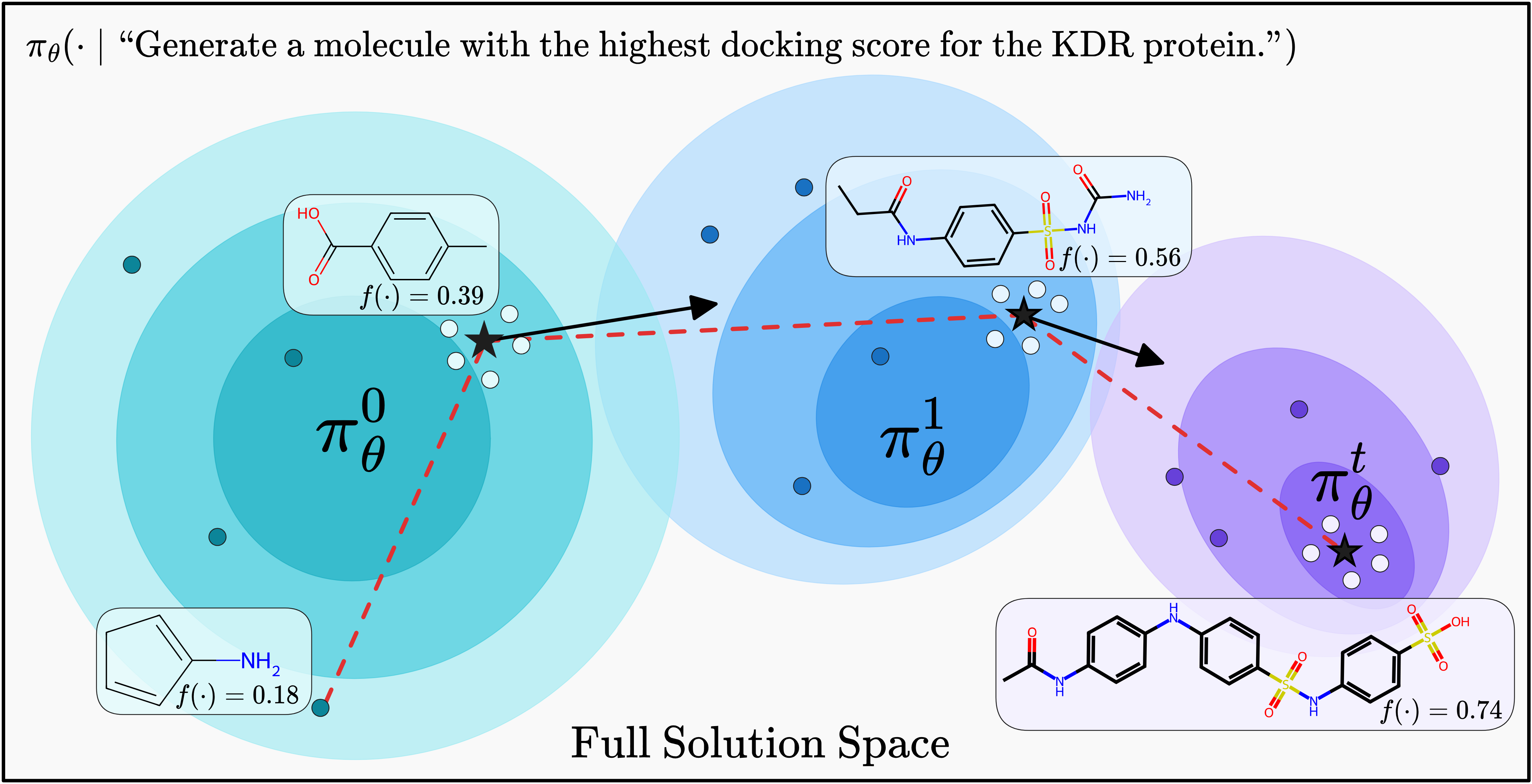}
    \caption{\textbf{Overview of \method{}.} Given a search problem, \method{} iteratively searches for optimal solutions by sampling candidates and updating its policy model $\pi_\theta^t$ using mixed-policy GRPO.
    In each iteration, we combine online samples ($\bullet$) from the current policy distribution,
    top-performing past solutions ($\star$) as greedy references,
    and samples drawn from the neighborhoods of greedy solutions ($\circ$) to form a GRPO group. 
    The resulting group is used to update $\pi_\theta^t$ and \emph{migrate} towards a sampling distribution that is likely to generate higher-quality solutions according to $f(\cdot)$.
    }
    \label{fig:teaser}
\end{figure}

Large language models (LLMs) have emerged as general-purpose tools for solving a wide range of black-box optimization problems \cite{boiko2023autonomous, ramos2023bayesian, liu2024large}. These models offer a flexible interface for generating candidate solutions, both in structured tasks, e.g., molecule design \cite{rankovic2023bochemian, pmlr-v235-kristiadi24a, gruver2024finetuned}, and unstructured, natural-language tasks, e.g., scientific hypothesis generation \cite{lu2024ai, majumder2025discoverybench, agarwal2025open}.

Recent work has shown that in-context learning (ICL) \cite{brown2020language} can effectively be used to steer LLMs toward higher-quality outputs in such tasks \cite{meyerson2023language, yang2024large, agarwal2025searching}. 
However, ICL alone lacks a principled mechanism to balance \emph{exploration} of novel solution areas with \emph{exploitation} of known high-reward ones \cite{krishnamurthy2024can} based on simply injecting a history of candidates in-context. 
Without this balance, the model may either get trapped in local optima or waste sampling budget on unpromising regions of the solution space. 

To improve LLM-based search, recent methods have explored \emph{test-time training} (TTT) \cite{sun2020test, hardt2024testtime}---a paradigm inspired from the human ability to generalize from a few examples \cite{yu2025learning}, in which the LLM is adapted at inference time for a specific problem instance before sampling a set of candidate solutions to evaluate. 
Similarly, some works have explored the use of off-policy reinforcement learning to efficiently learn suitable sampling distributions \cite{levine2020offline, yan2025learning}.
However, these approaches either rely on carefully hand-crafted, task-specific data generation strategies or assume availability of expert demonstration data \cite{akyürek2025surprisingeffectivenesstesttimetraining, li2024combininginductiontransductionabstract}, both of which limit the generality and scalability of such solutions.

To address these shortcomings, we cast search as an online reinforcement learning problem and leverage group relative policy optimization (GRPO) \cite{shao2024deepseekmath} to iteratively 
find promising regions of the search space, balancing exploration and exploitation. 
We, thus, propose \textbf{\method{}} (\textbf{Mi}xed-policy \textbf{GR}PO for \textbf{A}daptation at \textbf{Te}st-Time), a method for \emph{online} TTT that enables adaptive search with LLMs \emph{without} requiring any external, handcrafted training data\footnote{Our code is available at: \url{https://github.com/dhdhagar/migrate}.}. 
Our method combines:
\begin{enumerate}[left=1em]
    \item \textbf{On-policy sampling}, which ensures continual exploration of the solution space,
    \item \textbf{Greedy sampling}, which reuses top-performing past completions to exploit known high-reward regions, and
    \item \textbf{Neighborhood sampling (NS)}, which generates structurally similar variants of high-reward completions to facilitate local exploration.
\end{enumerate}

Crucially, all components in \method{} use only model-generated signals, eliminating the need for any external training data. We perform experiments on three challenging domains with diverse solution spaces and reward functions---word search, molecule optimization, and hypothesis+program induction using the challenging Abstraction and Reasoning Corpus (ARC) \cite{chollet2019measure}.
Across all domains, \method{} consistently outperforms both inference-only and TTT baselines, demonstrating that online TTT with mixed-policy guidance offers a scalable and general approach to LLM-based black-box optimization.

To summarize, our main contributions are as follows:
\begin{itemize}[left=1em]
    \item We introduce \method{}, a method to search for optimal solutions with LLMs using an online test-time training (TTT) algorithm without external demonstrations.
    \item We propose a mixed-policy group construction strategy that combines on-policy sampling with two novel off-policy techniques---greedy sampling and neighborhood sampling.
    \item We conduct comprehensive experiments across three diverse domains, showing that \method{} outperforms both inference-only and TTT baselines in complex black-box optimization tasks.
\end{itemize}

\section{Related Work}
\label{sec:related-work}

\paragraph{Test-time training.} 
Test-time training (TTT) aims to improve model performance on distribution shifts by updating models at inference. 
\citet{sun2020test} introduced TTT using a self-supervised objective on images to adapt network weights at test time. 
\citet{hardt2024testtime} demonstrate that fine-tuning LLMs on data closely related to each test prompt can yield large accuracy gains, extending TTT to reasoning tasks.
\citet{hubotter2025efficiently} show that nearest-neighbor retrieval for test-time fine-tuning often wastes effort on redundant examples, and instead propose an active-learning method that chooses maximally informative examples to reduce model uncertainty.

\paragraph{Local-structure methods.}
Instance-based learning (or ``local learning'') \cite{atkeson1997locally} is a common framework in machine learning where local structure is exploited around a test point to improve model accuracy, e.g., locally-weighted regression \cite{cleveland1979robust}.
In modern practice, this manifests as retrieving nearest-neighbor examples to guide adaptation, referred to as retrieval-augmented generation (RAG) or case-based reasoning (CBR) \cite{lewis2020retrieval, das-etal-2021-case, thai-etal-2023-machine, agarwal-etal-2024-bring}. 
In reinforcement learning, local policy search methods (e.g., off-policy local improvements, trust-region updates) behave like hill-climbers in the policy space.

\paragraph{Evolutionary computation.}
EvoTune \cite{surina2025algorithm} uses an LLM as a policy-generating operator in an evolutionary loop, then applies RL fine-tuning to iteratively improve it.
AlphaEvolve \cite{novikov2025alphaevolve} similarly creates an agent that uses multiple LLMs and automated evaluators to propose and refine codebases via an evolutionary framework.
FunSearch \cite{romera2024mathematical} pairs a pre-trained LLM with an automated evaluator and repeatedly samples and scores code functions, effectively evolving programs to solve mathematical problems.
In these systems, the ``population'' of programs or policies evolves over generations, often via an islands model or parallel ensembles, to avoid local traps.

\paragraph{RLVR.}
Reinforcement Learning with Verifiable Rewards (RLVR) \cite{lambert2025tulu3pushingfrontiers, deepseekai2025deepseekr1incentivizingreasoningcapability} is an approach for fine-tuning LLMs using RL guided by ground-truth reward functions, in contrast to typical RL-based methods that rely on learned or heuristic-based reward functions, which can introduce ambiguity.
In mathematics and code generation, these rewards are determined by correctness, such as matching a ground-truth solution or passing unit tests \cite{lambert2025tulu3pushingfrontiers, deepseekai2025deepseekr1incentivizingreasoningcapability, kimiteam2025kimik15scalingreinforcement}. Recently, RLVR has been instrumental in developing reasoning-based LLMs such as OpenAI-o1 \cite{openai2024openaio1card} and DeepSeek-R1 \cite{deepseekai2025deepseekr1incentivizingreasoningcapability}.

\section{Background}
\label{sec:background}

\paragraph{GRPO.} Group relative policy optimization \cite{shao2024deepseekmath} is a reinforcement learning algorithm used
to fine-tune LLMs
that replaces the value function in PPO training \cite{Schulman2017ProximalPO} with an estimate derived from Monte Carlo samples instead.
In particular, in each iteration of training, GRPO constructs a group \group{} of \groupsize{} completions, typically sampled from the current model, and calculates the advantage for every completion as a relative comparison to the group.
Let $\pi_{\theta_{\text{old}}}$ and $\pi_\theta$ denote the model policies (LLM parameters, in our case) before and after taking a gradient step. Given a task prompt \taskprompt{} and a set of completions sampled from the current model $\{o_i : o_i \sim \pi_{\theta_\text{old}} \}_{i=1}^N$, the GRPO loss objective is defined as

\begin{align}
    \mathcal{L}_{\text{GRPO}}(\theta) = 
    & - \frac{1}{\sum_{i=1}^N |o_i|} 
    \sum_{i=1}^N \sum_{t=1}^{|o_i|} \Big[
    \min\big(r_{i,t}(\theta)\hat{A}_{i,t}, \mathrm{clip}(r_{i,t}(\theta), 1 - \varepsilon_\text{low}, 1 + \varepsilon_\text{high}) \hat{A}_{i,t} \big)
    \Big] \label{eq:grpo}
\end{align}

where
\begin{align}
r_{i, t}(\theta) &= 
\frac{\pi_\theta(o_{i, t} \mid \taskprompt, o_{i, < t})}
     {\pi_{\theta_{\text{old}}}(o_{i, t} \mid \taskprompt, o_{i, < t})}, \quad\quad\quad \hat{A}_{i, t} = r_i - \mathrm{mean}(\{f(o_i)\}_{i=1}^N) \notag
\end{align}

are the policy ratio and advantage estimates, respectively, for each token in each completion, $f(\cdot)$ is a reward function that provides a scalar score for each completion, $\mathrm{clip}(\cdot,\cdot,\cdot)$ is a clipping function to prevent large updates during optimization, and $\varepsilon_\text{low/high}$ are clipping hyperparameters.

\paragraph{On-, off-, and mixed-policy optimization.} Typically, reinforcement learning (including GRPO) operates in an \emph{on-policy} manner, where new solutions are sampled using $\pi_\theta$ (i.e., the policy being trained) to estimate the loss for the next training step. On the other hand, some works have argued that on-policy training may constrain learning to only the capabilities of the base LLM itself, resulting in echo chambers \cite{zhao2025echo, yue2025does} that prevent novel task generalization. This problem is further exacerbated in the sparse reward scenario, where the base model is unable to generate solutions that elicit non-zero reward, thus leading to degenerate policy gradients. 
To address this, \emph{off-policy} optimization \cite{levine2020offline} has been proposed as an effective strategy that leverages previously collected expert demonstrations for training instead of online samples. However, a purely offline strategy can result in learning policies that are unable to generalize at inference time \cite{fujimoto2019off, kumar2019stabilizing}. 
Consequently, recent work \cite{yan2025learning} shows that a combination of online and offline samples, called \emph{mixed-policy} optimization, can outperform either strategy used in isolation.

\section{\method{}: Methodology}
\label{sec:method}

The focus in this work is on finding optimal solutions with respect to a black-box objective function $f(\cdot)$ under a finite sampling budget \budget.
To this end, we are interested in using GRPO as a \emph{search} algorithm, wherein a single example query is used as the input for a search task across multiple sampling iterations. The goal, then, is to learn query-specific parameters that shift the model's sampling distribution iteratively, improving the quality of solutions that are generated.\footnote{This is in contrast to the more typical setting of training a generalizable model with multiple examples. See the appendix for a complete description of modifications we incorporate from previous work beyond the original formulation from \citet{shao2024deepseekmath}.} Note that throughout this work, we use LoRA fine-tuning \cite{hu2022lora} instead of full-model training.

\paragraph{Overcoming sparse rewards in search.} As described earlier, purely on-policy learning is often unable to find an appropriate sampling distribution for a single query within a limited budget due to sparse rewards, i.e., when solutions sampled from the current policy do not result in useful policy gradients to make progress. At the same time, both off- and mixed-policy strategies require access to known expert demonstrations, which we assume are not available in our setting. We, therefore, present \textbf{\method{}}---a mixed-policy optimization strategy for GRPO that generates off-policy data via (a) selecting high-performing solutions from the model's own sampling history, and (b) sampling variations from the neighborhoods of observed high-performing solutions. 
In each iteration, \method{} combines new on-policy and off-policy samples to construct a group of completions \group{}, which is used to compute a policy gradient with respect to the loss function in Equation~\ref{eq:grpo}, until either the optimal solution is found or the sampling budget is exhausted.

\subsection{Mixed-Policy Group Construction for Search}
\label{subsec:group-construction}
Given a search task \task{} and a corresponding task prompt \taskprompt{} for the LLM, our goal is to construct a new group $\group_t$ composed of \groupsize{} completions in each search iteration $t$ to compute a policy gradient via GRPO.
We introduce two off-policy data selection techniques---\textbf{greedy} and \textbf{neighborhood sampling (NS)}---which we combine with on-policy sampling to generate test-time training data. 
Intuitively, both techniques are designed to bias policy gradients to \emph{exploit} known high-quality solutions sampled thus far, while on-policy sampling encourages \emph{exploration}.
Note that for a single iteration, we limit the number of new completions sampled from the LLM, regardless of policy, to \groupsize{}.
In experiments, we find that the simultaneous application of greedy and NS off-policy data selection (i.e., \method; Algorithm~\ref{alg:algorithm}) results in the best performance.

\paragraph{On-policy sampling.} Let \nonline{} ($\le \groupsize$) be the number of completions sampled from the current policy model, i.e., at timestep $t$, we generate on-policy completions (or observations) $\obsonline := \{o_i : o_i \sim \pi_\theta^{t-1}(\cdot \mid \taskprompt)\}_{i=1}^\nonline$ using temperature-based ancestral sampling.

\paragraph{Greedy sampling.} 
Let \database{} be a database of completions, which may be composed both of any candidate solutions available \textit{a priori} as well as all attempts sampled from the model in previous search iterations. 
In greedy off-policy data selection, if $\database \ne \emptyset$, we sample \ngreedy{} ($\le \groupsize$) known completions from $\database$ that are high-quality.
In particular, we first greedily select the top-$k$ completions from $\database$ with respect to $f(\cdot)$ and then randomly sample \ngreedy{} completions from the top-$k$, i.e., $\obsgreedy := \{o_i : o_i \sim \mathrm{topk}_f(\database) \}_{i=1}^\ngreedy$, where $\mathrm{topk}_f(\database)$ returns the $k$ best completions from $\database$ with respect to $f$.

\begin{wrapfigure}{R}{0.55\textwidth}
\vspace{-1.5em}
\begin{minipage}{0.54\textwidth}
\begin{algorithm}[H]
\caption{\textbf{Solution search with \method}}
\label{alg:algorithm}
\textbf{Input}: Task \task, black-box function $f$, budget \budget \\
\textbf{Parameters}:
GRPO group size \groupsize, \nonline{} on-policy samples, \ngreedy{} greedy samples, \nnbr{} neighborhood samples \\
\textbf{Output}: Best solution $o_\text{best}$
\begin{algorithmic}[1]
\STATE \textbf{Initialize:} Policy $\pi_\theta^0 \leftarrow \texttt{LLM}$, task prompt \taskprompt, database $\mathcal{D} \leftarrow \emptyset$, timestep $t \leftarrow 0$, $o_\text{best} \leftarrow \emptyset$
\WHILE{$|\database| < \budget$}
    \STATE $t \leftarrow t + 1$
    \STATE $\obsonline \leftarrow \{o_i : o_i \sim \pi_\theta^{t-1}(\cdot \mid \taskprompt)\}_{i=1}^\nonline$ %
    \STATE $\obsgreedy \leftarrow \{o_i : o_i \sim \mathrm{topk}_f(\database) \}_{i=1}^\ngreedy$
    \STATE $\nsprompt \leftarrow$ Build NS prompt using \obsgreedy
    \STATE $\obsns \leftarrow \{o_i : o_i \sim \pi_\theta^{t-1}(\cdot \mid \nsprompt)\}_{i=1}^\gamma$
    \STATE $\group_t \leftarrow \obsonline \oplus \obsgreedy \oplus \obsns$
    \STATE $\database \leftarrow \database \oplus \obsonline \oplus \obsns$
    \STATE $o_\text{best} \leftarrow \argmax_{o_i \in \database}f(o_i)$
    \IF{$o_\text{best}$ is optimal}
        \RETURN $o_\text{best}$
    \ENDIF
    \STATE $\pi_\theta^t \leftarrow$ Update using GRPO with $\group_t$ (Eq.~\ref{eq:grpo})
\ENDWHILE
\RETURN $o_\text{best}$
\end{algorithmic}
\end{algorithm}
\end{minipage}
\end{wrapfigure}

\paragraph{Neighborhood sampling.} 
While greedy sampling explicitly encourages the exploitation of high-quality samples, it is limited to leveraging solutions that have already been generated and may be prone to optimizing for local optima \cite{krishnamurthy2024can, agarwal2025searching}. To mitigate this, we incorporate a complementary off-policy sampling strategy grounded in 
a \emph{continuity assumption}---namely, that small variations in solutions yield small changes in quality. 
This assumption motivates exploration within neighborhoods of known high-quality candidates by prompting the model to generate stochastic variations of greedy samples, thereby producing \emph{new} solutions that may both provide useful variations for better policy gradients as well as solutions that may outperform previous samples. 
In practice, we construct a single neighborhood sampling prompt \nsprompt{} composed of all \ngreedy{} greedy samples along with an instruction to generate \nnbr{} ($\le \groupsize$) solution variations to construct $\obsns := \{o_i : o_i \sim \pi_\theta^{t-1}(\cdot \mid \nsprompt)\}_{i=1}^\gamma$.

\paragraph{\method.}
To balance exploration and exploitation during test-time training with GRPO, \method{} integrates both off-policy techniques with on-policy sampling by combining $\obsonline$, $\obsgreedy$, and $\obsns$ into a single group $\group_t$, with the constraint that $\nonline + \nnbr <= \groupsize$ in each iteration\footnote{We keep constant the number of new solutions sampled from the LLM for fair comparison with baselines. In practice, we ensure that $\nonline + \ngreedy + \nnbr = \groupsize$ to simplify our implementation.} (see Algorithm~\ref{alg:algorithm}). We compute the loss on $\group_t$ with respect to the task prompt \taskprompt, irrespective of how the sample was generated.
While on-policy sampling encourages exploration of new solutions, greedy sampling promotes exploitation by reusing high-quality completions from a running database, and neighborhood sampling introduces structured exploration via local variations of the greedy samples. Empirically, we find that this combination produces higher-quality search results than any single strategy alone.

\section{Experiments}
\label{sec:experiments}

\subsection{Search Tasks}
Following \citeauthor{agarwal2025searching}, we evaluate \method{} by conducting experiments on three text-based search tasks---Semantle (word search), Dockstring (molecule optimization), and ARC (hypothesis+program search).

\paragraph{Semantle.}
Semantle \cite{agarwal2025searching} is a word-search task, where the goal is to identify a held-out English word (e.g., \textit{``polyethylene''}) within a limited number of guesses. The black-box function used indicates how semantically close a guessed word is to the target, which is computed using cosine similarities over SimCSE \cite{gao2021simcse} embeddings, following prior work. Each search problem is initialized with a warmstart set of 20 words (randomly sampled from the word2vec index \cite{mikolov2013efficient}) and corresponding black-box scores. We conduct evaluation using 10 hidden words and 5 warmstart sets for each of them, resulting in a total of 50 problem instances.

\paragraph{Dockstring.}
\citeauthor{garcia2022dockstring} provides a suite of challenging molecule optimization tasks that reflect real-world problems in drug discovery. We focus on a multi-objective optimization task: generating molecules (represented as SMILES strings \cite{weininger1988smiles}) that simultaneously maximize druglikeness and binding affinity, quantified by QED \cite{bickerton2012quantifying} and negative Vina scores \cite{trott2010autodock}, respectively. We use a scalarized multi-objective black-box function (Equation~\ref{eq:dockstring}) that places a greater weight on Vina scores than QED, reflecting the common prioritization of binding affinity over druglikeness when evaluating a molecule's drug efficacy \cite{https://doi.org/10.1111/j.1476-5381.2010.01127.x, Wenlock2003-jm}. Following prior works \cite{yuksekgonul2024textgrad, agarwal2025searching}, we run evaluations for 58 pharmaceutically-relevant protein targets.

\paragraph{ARC.}
The Abstraction and Reasoning Corpus (ARC) \cite{chollet2019measure} is a benchmark of grid-based puzzles that involves inferring the transformation logic from a small set of input-output grid pairs and applying it to a held-out test grid. 
Recent methods improve performance via data augmentation with invertible transformations \cite{akyürek2025surprisingeffectivenesstesttimetraining} or by combining program synthesis with transductive strategies \cite{li2024combininginductiontransductionabstract}. 
We take an inductive hypothesis + program search approach \cite{wang2024hypothesis}, where natural language transformation algorithms are hypothesized and translated into Python programs.
We report two accuracy metrics: \emph{pass@2}, which measures whether any of the top-2 common outputs from the programs that solve the train set matches the test grid, and \emph{oracle}, which 
provides credit if any of the sampled programs correctly solves the test grid. 
Note that oracle accuracy reflects a coarse ability to find a distribution that can generate the correct solution.

We conduct our experiments on two dataset versions: \textbf{ARC-Full} and \textbf{ARC-Small}. ARC-Full includes all 400 tasks from the ARC evaluation set \cite{arc-prize-2024}, while ARC-Small is a subset consisting of 54 tasks with grids up to a maximum of 64 cells. We create this small subset to measure variance across search methods via repeat runs.
Note that we ensure ARC-Small maintains the same difficulty distribution as ARC-Full\footnote{Due to hardware limitations, we truncate prompts at 2048 tokens in all experiments. As a result, only 200 out of 400 tasks in ARC-Full could be evaluated with their full context.}. To guide search, we follow prior work \cite{agarwal2025searching} and use a Hamming-distance based black-box function. %

\subsection{Baselines}

\paragraph{Inference-only.} We evaluate three inference-only sampling strategies for optimization tasks:  
\begin{itemize}[left=1em]
    \item \textbf{Random}, which generates completions by sampling directly from the base model using only the task prompt;
    \item \textbf{Neighborhood Sampling (NS)}, which samples completions from a prompt that includes top-performing solutions from previous iterations to encourage local exploration; and
    \item \textbf{OPRO} \cite{yang2024large}, which generates completions using a prompt that builds a trajectory of top-performing solutions as a textual gradient to discover new solutions that may improve performance.
\end{itemize}

\paragraph{Test-time training.} Beyond inference-only methods, we evaluate three variants of our GRPO-based test-time training approach:
\begin{itemize}[left=1em]
    \item \textbf{GRPO} is the base algorithm, using a fixed task prompt and sampling $\groupsize$ completions on-policy from the current model (i.e., $\nonline = \groupsize$, $\ngreedy = 0$, $\nnbr = 0$).
    \item \textbf{GRPO-Greedy} augments GRPO by using greedy off-policy sampling to select \ngreedy{} previous completions to place in the group at each iteration (i.e., $\nonline, \ngreedy > 0$ and $\nnbr = 0$).
    \item \textbf{\method{}} is our full method, combining on-policy exploration, greedy sampling of top completions, and neighborhood sampling for local exploration (i.e., each of $\nonline, \ngreedy, \nnbr > 0$).
\end{itemize}

We provide complete details of our experiment setting in the appendix, including the values used for \nonline{}, \ngreedy{}, and \nnbr{} for different tasks. We also provide a sensitivity analysis of these choices on the Semantle task in the Results section.

\paragraph{Additional baselines.} 
We also evaluate \method{} (OPRO), a variant of \method{} that replaces the neighborhood sampling (NS) prompt with the OPRO prompting strategy for local exploration. Additionally, we explore an alternative strategy for selecting $\obsgreedy$ using an islands-based evolutionary search method. Please see the appendix for both sets of results.

\paragraph{Models.} 
We present our main results on Semantle and Dockstring using Llama-3.2-3B-Instruct \cite{llama3modelcard}. For ARC, we use Llama-3.1-ARC-Potpourri-Induction-8B \cite{li2024combininginductiontransductionabstract}, a fine-tuned version of Llama-3.1-8B-Instruct \cite{llama3modelcard} trained on synthetic Python programs that solve ARC training tasks. The latter decision is driven by the bespoke nature of the ARC challenge, where base models are entirely unable to generate valid solutions.

\begin{figure*}[t]
    \centering
    \subfloat[Semantle]{%
        \includegraphics[width=0.24\textwidth]{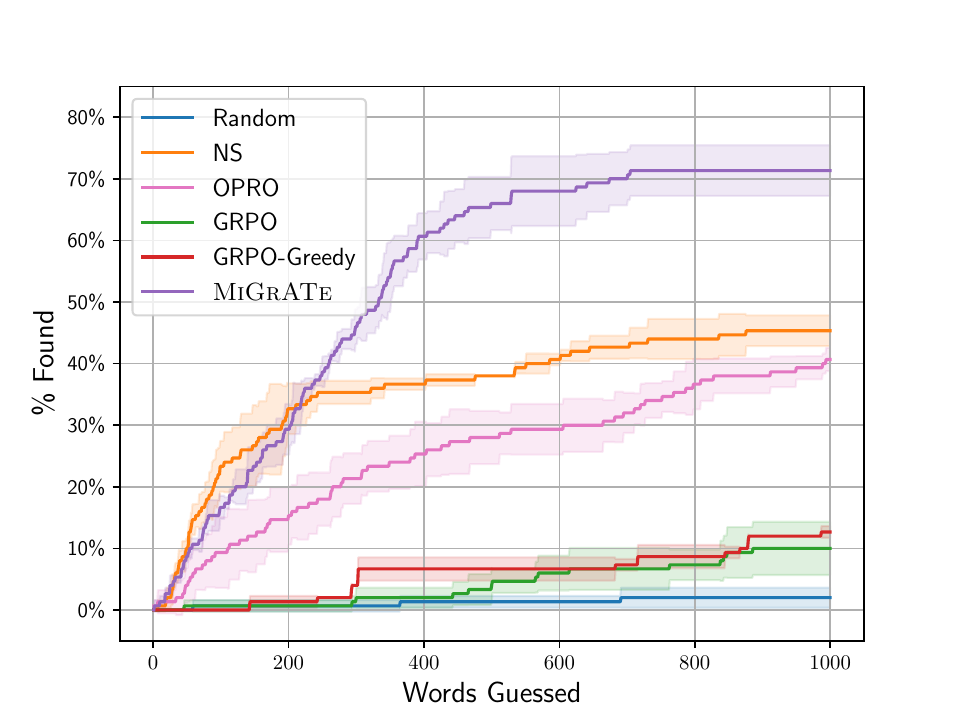}
    }
    \hfill
    \subfloat[Dockstring]{%
        \includegraphics[width=0.24\textwidth]{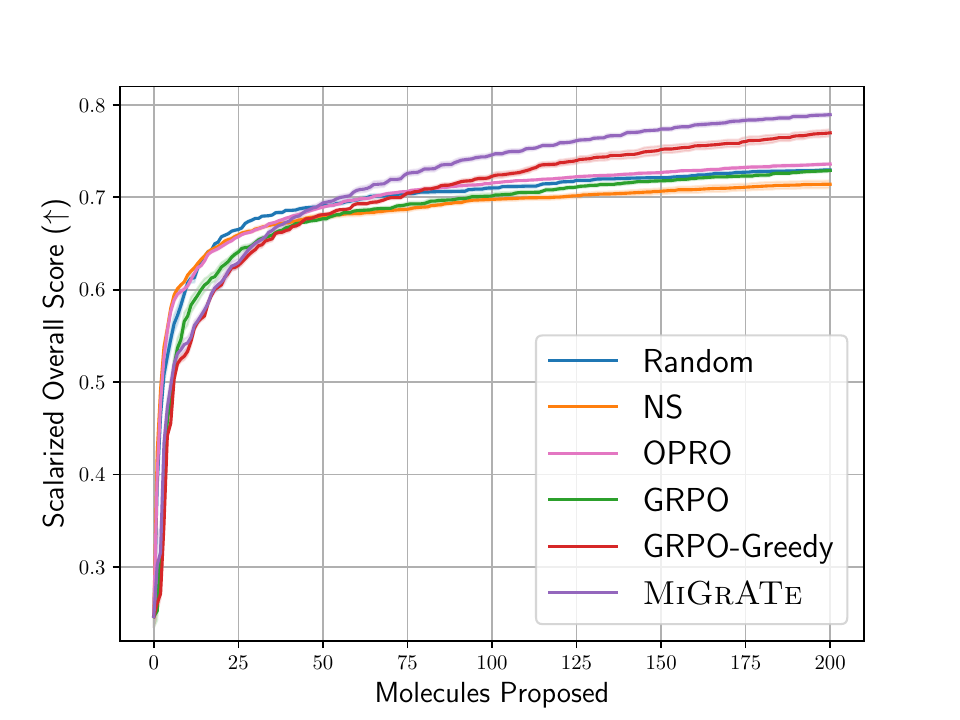}
    }
    \hfill
    \subfloat[ARC-Small]{%
        \includegraphics[width=0.24\textwidth]{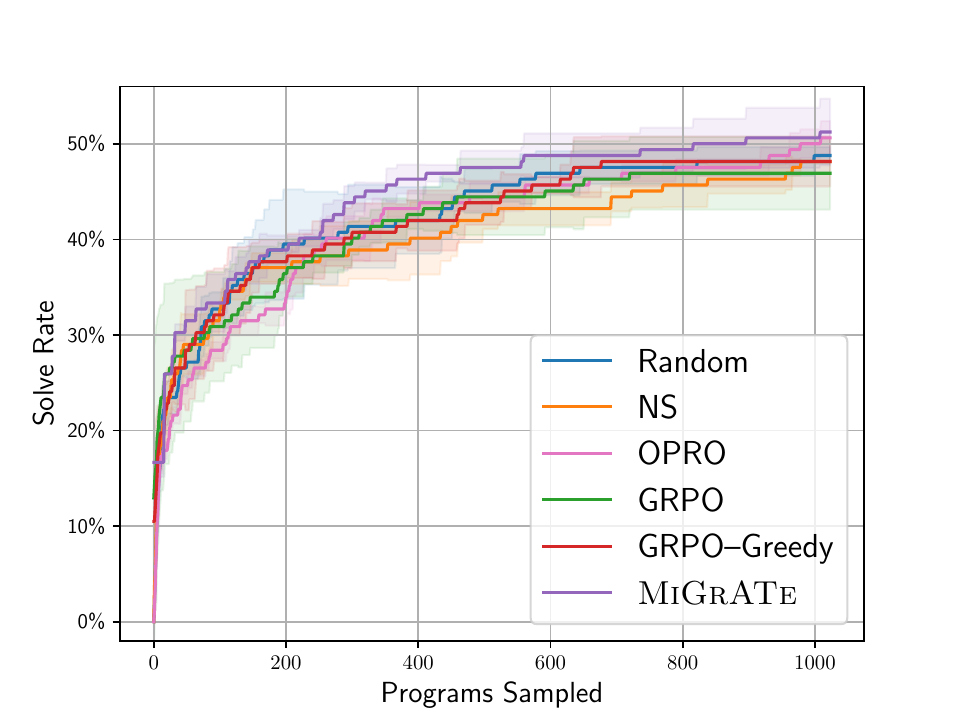}
    }
    \hfill
    \subfloat[ARC-Full]{%
        \includegraphics[width=0.24\textwidth]{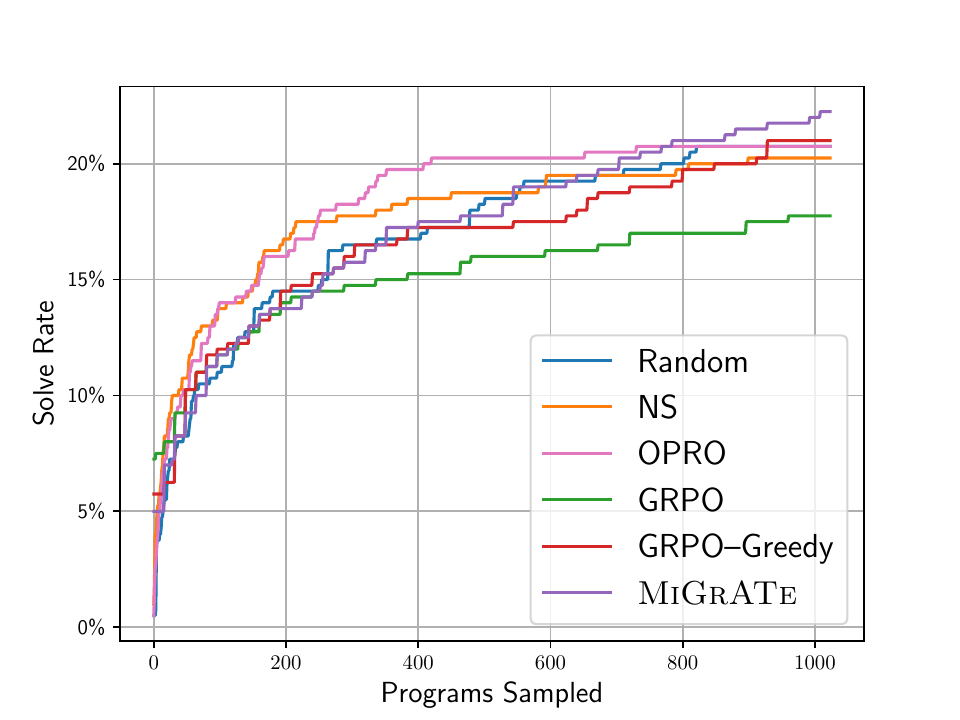}
    }

    \caption{\textbf{Best-so-far performance results.} (a) On Semantle, \method{} outperforms all baselines, improving the second-best (NS) by $25\%$. (b) In Dockstring, \method{} surpasses baselines after 50 proposals. (c) On ARC-Small, \method{} upper-bounds baselines across budget levels. (d) On ARC-Full, \method{} solves more tasks than baselines at the full budget.}
    \label{fig:performance_plots}
\end{figure*}

\section{Results and Discussion}
\label{sec:results}

\paragraph{\method{} outperforms both inference-only and TTT baselines.} Across tasks, we run each method until either the correct solution is found or a pre-defined budget of solution candidates (1000 for Semantle, 200 for Dockstring, and 1024 for ARC) is proposed and evaluated.\footnote{We report all configuration parameters for \method{} and other baselines in the appendix.}
We report our results on each search task in Table~\ref{tab:combined-table} and provide a best-so-far plot to trace search behavior across sampling budgets in Figure~\ref{fig:performance_plots}. 
We find that mixed-policy GRPO via \method{} outperforms each inference-only baseline as well as the TTT-based ablations. 

In Semantle, our results show that \method{} outperforms baselines by $\ge25$ percentage points. 
Notably, as shown in Figure~\ref{fig:performance_plots}a, across the 50 problem instances averaged over 3 repeat runs, 
\method{} surpasses its inference-only counterpart NS after 200 guesses ($\sim$20 \method{} iterations), demonstrating the benefit of performing explicit gradient updates in finding sampling distributions with higher-quality solutions versus using a purely in-context optimization strategy.

\begin{table*}[hpt]
  \centering
  \scriptsize
  \begin{tabular}{l c ccc} \toprule
    & \multicolumn{1}{c}{\textbf{Semantle}} & \multicolumn{3}{c}{\textbf{Dockstring}}\\
    \cmidrule(lr){2-2}
    \cmidrule(lr){3-5}
    \textbf{Method} & \textbf{\% Found} & \textbf{QED ($\uparrow$)} & \textbf{Vina Score ($\downarrow$)} & \textbf{Overall Score ($\uparrow$)}\\
    \midrule
    Random & $2.00 \pm 1.63$ & \bestresult{0.91 \pm 0.00} & $-9.92 \pm 0.15$ & $0.73 \pm 0.00$\\
    NS & \secondbestresult{45.30 \pm 2.49} & $0.87 \pm 0.01$ & $-9.65 \pm 0.21$ & $0.71 \pm 0.00$\\
    OPRO & $40.70 \pm 1.89$ & $0.90 \pm 0.00$ & $-9.94 \pm 0.06$ & $0.74 \pm 0.00$\\
    GRPO & $10.00 \pm 4.32$ & \secondbestresult{0.91 \pm 0.00} & $-10.09 \pm 0.05$ & $0.73 \pm 0.00$\\
    GRPO-Greedy & $12.70 \pm 0.94$ & $0.90 \pm 0.01$ & \secondbestresult{-10.80 \pm 0.19} & \secondbestresult{0.77 \pm 0.00}\\
    \midrule
    \method{} & \bestresult{71.30 \pm 4.11} & $0.90 \pm 0.00$ & \bestresult{-11.00 \pm 0.07} & \bestresult{0.79 \pm 0.00} \\
    \bottomrule
  \end{tabular}
  
  \vspace{1em}
  
  \begin{tabular}{l cc cc}
    \toprule
    & \multicolumn{2}{c}{\textbf{ARC-Small}} & \multicolumn{2}{c}{\textbf{ARC-Full}}\\
    \cmidrule(lr){2-3}
    \cmidrule(lr){4-5}
    \textbf{Method} & \textbf{Pass@2 (\%)} & \textbf{Oracle (\%)} & \textbf{Pass@2 (\%)} & \textbf{Oracle (\%)} \\
    \midrule
    Random & $48.20 \pm 1.51$ & $57.41 \pm 0.87$ & $20.75$ & $28.00$ \\
    NS & $48.15 \pm 0.00$ & $55.56 \pm 1.51$ & $20.25$ & \secondbestresult{29.50} \\
    OPRO & \secondbestresult{50.62 \pm 1.75} & \secondbestresult{59.26 \pm 0.00} & $20.75$ & $27.75$ \\
    GRPO & $46.91 \pm 3.81$ & $55.56 \pm 6.90$ & $17.75$ & $27.00$ \\
    GRPO-Greedy & $48.15 \pm 2.62$ & $56.17 \pm 7.26$ & \secondbestresult{21.00} & \bestresult{30.00} \\
    \midrule
    \method{} & \bestresult{51.23 \pm 3.49} & \bestresult{62.35 \pm 0.87} & \bestresult{22.25} & \bestresult{30.00} \\
    \bottomrule
  \end{tabular}
\caption{\textbf{Search performance.} Except ARC-Full, results are averaged over three random seeds with standard deviations reported. The top-2 results in each column are marked with bold and underline, respectively. \method{} outperforms on all but one metric.}
\label{tab:combined-table}
\end{table*}

\begin{figure*}[t]
    \centering
    \subfloat[Dockstring search trace]{%
        \includegraphics[width=0.3\textwidth]{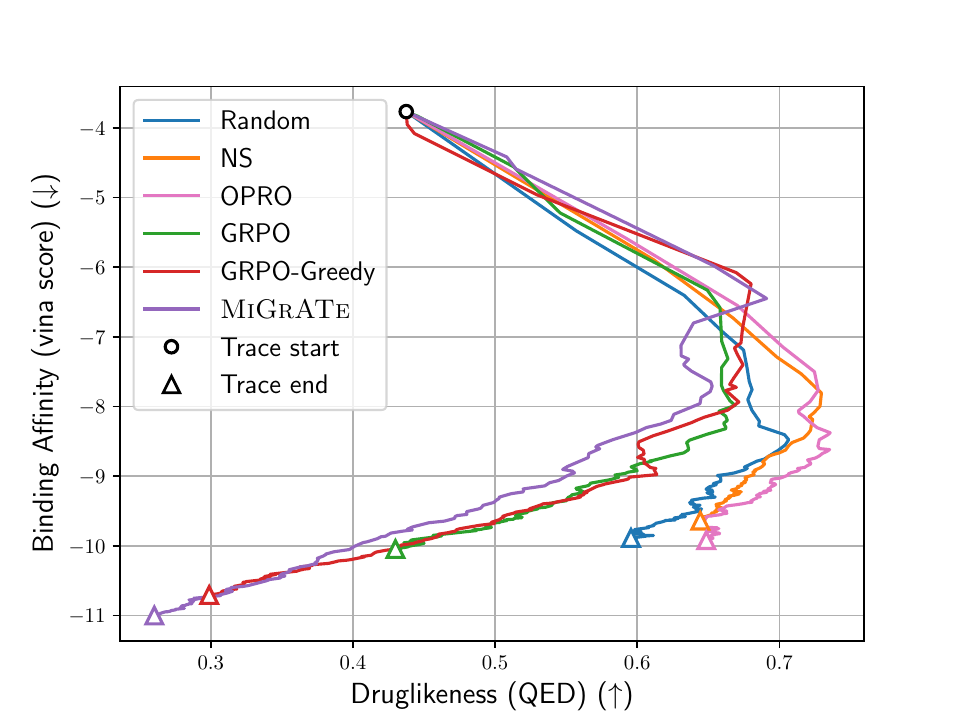}
        \label{fig:dockstring-trace}
    }
    \subfloat[SMILES distribution for KDR]{%
        \includegraphics[width=0.3\textwidth]{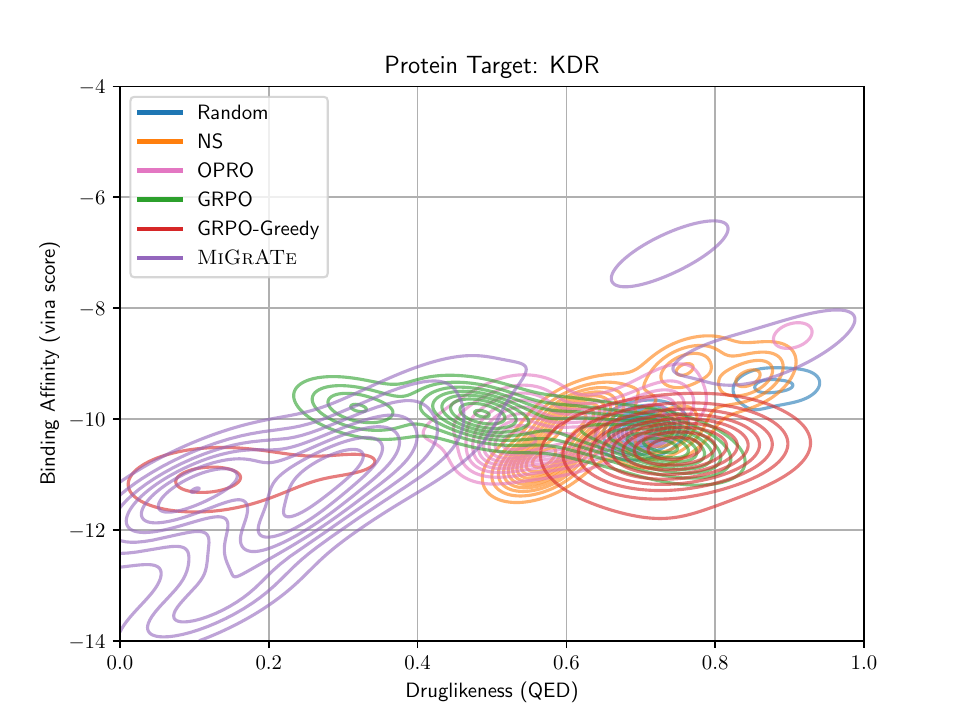}
        \label{fig:mol-contour-kdr}
    }

    \caption{\textbf{Dockstring search behavior.} (a) Vina and QED scores for best molecules found as search progresses. Each trace starts from 3 diverse fragments (acetamide, pentane, and benzene). (b) Distribution of binding affinity and druglikeness for KDR target. \method{} explores a broader region of chemical space, including low-affinity, low-druglikeness areas ignored by baselines.}
    \label{fig:dockstring-combined}
\end{figure*}

In Dockstring, 
we allocate a budget of 200 molecule proposals for each method and report performance over 3 repeat runs. Table~\ref{tab:combined-table} shows that \method{} synthesizes molecules with higher scalarized scores (according to Equation~\ref{eq:dockstring}), i.e., jointly optimizing for QED and Vina. Further, in Figure~\ref{fig:performance_plots}b, we see that \method{} outperforms all baselines on average after 50 molecule proposals. Additionally, we show the search trace of different methods in Figures~\ref{fig:dockstring-trace} and \ref{fig:mol-contour-kdr}.

In ARC, we report performance over 3 repeat runs on ARC-Small and a single run on ARC-Full due to hardware constraints. For each run, we allocate a search budget of 1024 programs.
From Figure~\ref{fig:performance_plots}c, Figure~\ref{fig:performance_plots}d, and Table~\ref{tab:combined-table}, we find that \method{} does outperform baselines but demonstrates more modest improvement.
We further find that \method{} solves all but two tasks that the baselines also solve.

\paragraph{TTT methods produce qualitatively different solutions than inference-only methods.}

In Semantle, across runs, we find that \method{} is the only method that is able to find all 10 hidden words. Furthermore, we observe that only \method{} and its ablations are able to optimize for certain words, e.g., ``birthstone'', indicating an ability to effectively navigate the unique search landscape for this word. 
In Dockstring, as shown in Figure~\ref{fig:dockstring-trace}, we find that the optimization  trajectories of the best-performing SMILES strings found using TTT methods (\method{} and its ablations) show a distinct pattern that optimize for Vina scores more heavily than those from inference-only methods, which prefer higher QED and are unable to synthesize molecules with lower than $-10$ kcal/mol Vina.
While \method{} is indeed capable of generating molecules with high QED scores ($>0.8$), optimization prefers to reduce QED to below $0.3$ in exchange for better Vina scores.  
This also follows from the scalarized multi-objective function in Equation~\ref{eq:dockstring}, which attaches a stronger weight to Vina scores than QED.

\begin{figure*}[t]
    \centering
    \subfloat[Semantle]{%
        \includegraphics[width=0.3\textwidth]{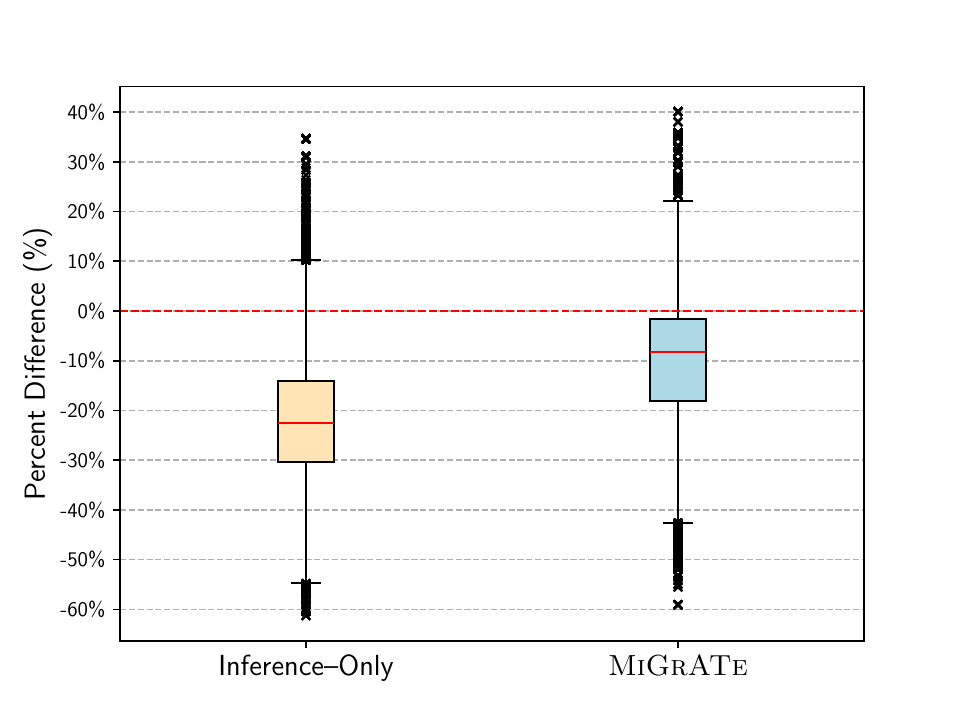}
    }
    \hfill
    \subfloat[Dockstring]{%
        \includegraphics[width=0.3\textwidth]{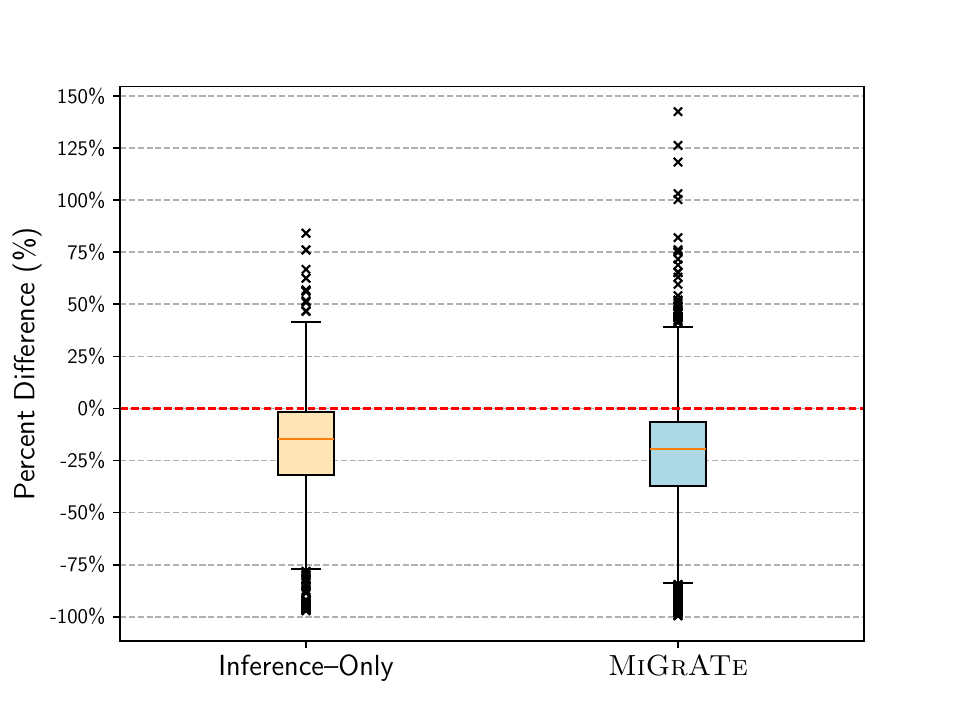}
    }
    \hfill
    \subfloat[ARC]{%
        \includegraphics[width=0.3\textwidth]{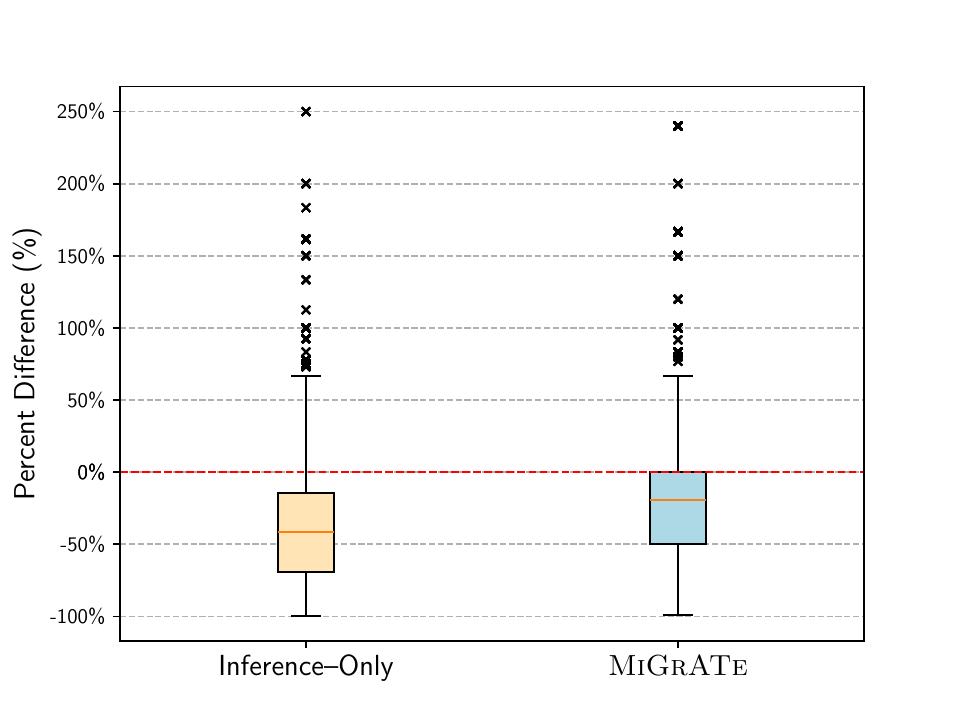}
    }
    \caption{\textbf{Performance relative to the best-so-far.} Percentage difference between samples from inference-only NS and \method{} with their best-so-far scores during optimization. \method{} produces more samples close to or above the best-so-far than NS. In Dockstring, despite \method{} producing more invalid molecules, its outliers show larger gains in performance than NS. Note that due to a high proportion of invalid molecules and programs, we omit samples with 0 rewards.}
    \label{fig:sample_shift}
\end{figure*}

\paragraph{What search behaviors are observed with \method{}?}
To understand this, we analyze the quality of samples generated by \method{} and compare them to those from the inference-only NS baseline in Figure~\ref{fig:sample_shift}.
More specifically, we measure the relative difference between the black-box score of each solution sampled by both methods and the best-so-far performance when that solution was sampled during optimization. We then compare the distributions of these differences between the two methods.
On Semantle and ARC, search with \method{} demonstrates the ability to iteratively improve upon its previously best-found solution in contrast to the behavior seen with the inference-only strategy, which often samples solutions that show no improvement. 
In Dockstring, on the other hand, we see \method{} produce a higher number of invalid molecules than inference-only approaches, indicating broader exploration of the solution space, as also shown in Figures~\ref{fig:dockstring-trace} and ~\ref{fig:mol-contour-kdr}. We find that many of the proposed molecules are longer and more complex SMILES strings, evidenced by a $44\%$ increase in average length. Despite proposing more invalid molecules, however, \method{} finds molecules that improve upon the best-so-far with larger performance improvements than with inference-only.

\paragraph{Sensitivity analysis of $\nonline,\ngreedy, \nnbr$.} In Figures~\ref{fig:vary_ns_plots} and \ref{fig:semantle_vary_beta}, we show how varying the number of online, greedy, and NS samples impacts search performance with \method{}. On Semantle, configurations using \textit{only} neighborhood samples or a large proportion of greedy samples outperform those with online samples, suggesting that off-policy variants can perform effective search. Dockstring benefits from having a mix of sample types, with the best results from a balanced configuration, indicating the need for a strategy that both explores and exploits the search space. In contrast, ARC-Small gives an example of a domain where a higher proportion of online samples is important for improved search. These results highlight both the flexibility of \method{} to apply different search strategies and the importance of tuning the mixed-policy composition of \method{} for each domain.\footnote{\textbf{Additional analyses:} (a) evaluation of whether TTT weights from solved tasks can help bootstrap search for related unsolved tasks; (b) performance of \method{} with NS swapped out for an alternative local structure sampling technique OPRO \cite{yang2024large} (see Appendix~\ref{appendix:additional_experiments}).}

\section{Conclusion}
\label{sec:conclusion}

We introduced \method{}, a method for online test-time training of LLMs that enables efficient search in black-box optimization tasks without requiring handcrafted training data. 
By leveraging Group Relative Policy Optimization (GRPO) along with a novel mixed-policy group construction strategy---comprising on-policy, greedy, and neighborhood sampling---\method{} effectively balances exploration and exploitation. 
Our experiments across three text-based domains demonstrate the efficacy of \method{} to improve LLM-based search. 
Future work may include scaling online TTT to multi-step decision-making and integrating stronger uncertainty-aware acquisition strategies to further improve sample efficiency.

\bibliography{iclr2026_conference}
\bibliographystyle{iclr2026_conference}

\appendix
\section{Appendix A}
\label{appendix:appendix_a}

\subsection{Experimental Settings}
\label{appendix:experimental_settings}

\textbf{Semantle.} The black-box function we use is the cosine similarity of vector representations generated using the SimCSE \cite{gao2021simcse} sentence embedding model, where the score for a proposed word $x$ for a hidden target word $y$ is computed by comparing the embeddings for the sequences "What is a \{x\}?" and "What is a \{y\}?". The number of warmstart candidates is 20. Our main results with NS and \method{} selects \obsgreedy{} by uniformly sampling among the top-3 completions found so far according to their black-box scores. 

In \method{}, we execute GRPO for 100 generation steps where we sample a batch of 10 words in each step for a total sampling budget of 1000 words. In each step, we sort the generated batch of words by their scores and construct a group of 5 completions, each consisting of 2 words each. Each completion is assigned the maximum score of the two words as its reward.

For the Random baseline, we sample 1000 words using the task prompt. For the NS baseline, we sample 10 words using the NS prompt for 100 iterations. Similarly, for the OPRO baseline, we also sample 10 words using the OPRO prompt for 100 iterations. We provide, in-context, the top-10 words found so far for every OPRO-based method.

\paragraph{Dockstring.} The black-box function we use is a linear function of the binding affinity (Vina) and druglikeness (QED). We use RDKit's $\mathtt{MolFromSmiles}$ to sanitize a given generated SMILES string. If this process fails due to an invalid format structure or molecule, we assign the generated molecule a score of 0. If the molecule is valid, we compute the QED and Vina scores on the given protein target. We then compute the overall score of these two metrics as follows:

\begin{align}
  & s_\text{overall}(\text{molecule, protein}) = 
  1 - \mathcal{N}(\mathtt{Vina}(\text{molecule, protein}) + (1 - \mathtt{QED}(\text{molecule}))
\label{eq:dockstring}
\end{align}

Where $\mathcal{N}$ denotes min-max normalization to the range [0,1]. The QED score is bounded between 0 and 1, and we assume the Vina score to be between 0 and -13.0 kcal/mol.
In practice, the binding affinity is a much higher priority than the druglikeness. Given our equation and the value ranges for computing $s_\text{overall}$, our black-fox function accurately emphasizes the Vina score about 10 times more than the QED score.

For the Random baseline, we sample 200 molecules using the task prompt. For the NS baseline, we sample 3 molecules using the task prompt and 2 molecules using the NS prompt in each iteration for 40 iterations. We select \obsgreedy{} from the top-1 molecule found so far in NS and \method{}. For the OPRO baseline, we sample 5 molecules using the OPRO prompt for 40 iterations. We provide, in-context, the top-5 molecules proposed so far for every OPRO-based method.

\begin{table}[t]
\centering
\begin{tabular}{ll}
\toprule
\textbf{Hyperparameter} & \textbf{Value} \\
\midrule
Model & Llama 3.2 3B Instruct \\ & \cite{grattafiori2024llama3herdmodels} \\
Learning rate & 1e-5 \\
Group size & 5 \\
LoRA rank & 64 \\ 
LoRA alpha & 16 \\
Training steps & 100 \\
Iterations per step & 2 \\
\midrule
GRPO [\nonline, \nnbr, \ngreedy] & $[5, 0, 0]$ \\
GRPO-Greedy [\nonline, \nnbr, \ngreedy] & $[4, 0, 1]$ \\
\method{} [\nonline, \nnbr, \ngreedy] & $[0, 4, 1]$ \\
\bottomrule
\end{tabular}
\vspace{0.5em}
\caption{\method{} hyperparameters for Semantle}
\label{tab:semantle-hyperparams}
\end{table}

\begin{table}[t]
\centering
\begin{tabular}{ll}
\toprule
\textbf{Hyperparameter} & \textbf{Value} \\
\midrule
Model & Llama 3.2 3B Instruct \\ & \cite{grattafiori2024llama3herdmodels} \\
Learning rate & 5e-5 \\
Group size & 5 \\
LoRA rank & 64 \\ 
LoRA alpha & 16 \\
Training steps & 40 \\
Iterations per step & 1 \\
\midrule
GRPO [\nonline, \nnbr, \ngreedy] & $[5, 0, 0]$ \\
GRPO-Greedy [\nonline, \nnbr, \ngreedy] & $[4, 0, 1]$ \\
\method{} [\nonline, \nnbr, \ngreedy] & $[2, 2, 1]$ \\
\bottomrule
\end{tabular}
\vspace{0.5em}
\caption{\method{} hyperparameters for Dockstring}
\label{tab:dockstring-hyperparams}
\end{table}

\paragraph{ARC.} The black-box function we use is a hamming-distance based metric. We run all input grids with the sampled program and compute the proportion of cells in the ground-truth grid that matches the output grid. We assign a reward of 0 if the program does not terminate within 10 seconds of execution. During training, the reward is given by averaging the score across all training input grids of the given ARC task. If the output grid is larger than the ground-truth, then we assign a score of 0.

For the Random baseline, we sample 1024 programs using the task prompt. For the NS baseline, we sample 12 programs using the task prompt and 4 programs using the NS prompt for 64 iterations. We note that this Random baseline is equivalent to the main evaluations ran by \citeauthor{li2024combininginductiontransductionabstract} Additionally, our TTT baselines on ARC in the inductive setting are not an entirely fair comparison to prior works that do TTT in the transductive setting. We select \obsgreedy{} as the top-1 program found so far for both NS and \method{}. Similarly, for the OPRO baseline, we sample 12 programs using the task prompt and 4 programs using the OPRO prompt for 64 iterations. Due to hardware limitations and to maintain a fair comparison with \method{}, we only provide one program in-context for the OPRO prompt.

\begin{table}[t]
\centering
\begin{tabular}{ll}
\toprule
\textbf{Hyperparameter} & \textbf{Value} \\
\midrule
Model & BARC \cite{li2024combininginductiontransductionabstract} \\
Learning rate & 1e-5 \\
Group size & 16 \\
LoRA rank & 128 \\ 
LoRA alpha & 32 \\
Training steps & 64 \\
Iterations per step & 1 \\
\midrule
GRPO [\nonline, \nnbr, \ngreedy] & $[16, 0, 0]$ \\
GRPO-Greedy [\nonline, \nnbr, \ngreedy] & $[15, 0, 1]$ \\
\method{} [\nonline, \nnbr, \ngreedy] & $[11, 4, 1]$ \\
\bottomrule
\end{tabular}
\vspace{0.5em}
\caption{\method{} hyperparameters for ARC}
\label{tab:arc-hyperparams}
\end{table}

\subsection{GRPO Formulation}
\label{appendix:grpo}

We remove the KL term in the original GRPO objective. Following DAPO \cite{yu2025dapoopensourcellmreinforcement}, we utilize token-level normalization, which assigns more balanced rewards to individually generated tokens---alleviating the bias towards longer responses. We also set $\varepsilon_\text{low}=0.2$ and $\varepsilon_\text{low}=0.28$ which DAPO finds to promote exploration of low-probability tokens that perform well. Dr. GRPO \cite{liu2025understandingr1zeroliketrainingcritical} also divides the sum of loss by a constant instead of the total sequence length to completely remove any completion length bias. Although we did not use this formulation in our experiments, there should be no substantial differences since there is not high variability in the solution lengths in the domains we studied. Following Dr. GRPO, we do not scale the advantage by the standard deviation of the group's rewards. By doing so, we avoid biasing weight optimization on groups that perform extremely well or poorly on a given prompt. While our online prompt always remains constant, this bias is relevant for our NS prompt which can vary across iterations.

\subsection{Computational Resources}
All experiments were conducted on a cluster of NVIDIA GPUs. We utilize a mixture of A100 (40GB and 80GB), L40S, and A40 GPUs. TTT methods on ARC-Full were only ran with A100 (80GB) GPUs due to the higher memory requirements.
Our implementation of \method{}
is based on the TRL 0.19.0 implementation of GRPO from Hugging Face \cite{vonwerra2022trl}. We also utilize Unsloth \cite{unsloth} and vLLM \cite{kwon2023efficient} to enable higher inferencing throughput and lower memory usage.
The average runtime for \method{} on each Semantle problem was 93 seconds on an A100 GPU. The average runtime for \method{} across all GPU types on each molecule optimization task was 7.5 minutes. The average runtime for \method{} on each ARC task with early stopping is 51 minutes on an A100 GPU.

\label{appendix:computation}

\section{Appendix B: Additional Experiments}
\label{appendix:additional_experiments}

\subsection{Island-based Evolution Algorithm}
\label{appendix:evolution_algorithm}

We implement an island-based evoluationary algorithm as an alterative to top-$k$ for selecting \obsgreedy. We created a database inspired by \cite{ellenberg2025generativemodelingmathematicaldiscovery} to store generated solutions and sample them for constructing neighborhood sampling. The island model organizes the solutions into isolated islands of solutions that are evolved independently.

At every training step, we iterate to another ``island'' in the database in a cyclic order. We then sample a solution stored at this island to construct our neighborhood sampling prompt. We note that unlike prior works \cite{ellenberg2025generativemodelingmathematicaldiscovery, surina2025algorithm} we do not construct additional subclusters of solutions within each island. This was done due to the low sampling constraints of our experiments but can also be seen as using a single cluster per island. Sampling from an island is carried out by an exploitation strategy with probability $p$ and an exploration strategy with probability $1 - p$. With the exploitation strategy, we randomly select a top solutions on the island that is also considered a globally top-$k$ solution across all islands. If the island does not have a solution that is in the top-$k$ solution for all islands then we fall back on the exploration strategy. With the exploration strategy, we randomly select among the top solutions on the island that are \textit{not} one of the globally top-$k$ solutions.

We periodically migrate a percentage of the top-performing solutions from each island to their neighboring islands according to a ring topology. This maintains a balance of exploring diverse solutions in isolation and preventing the algorithm from spending too much time on low-performing solutions.

\begin{figure}[htbp]
    \centering
    \includegraphics[width=0.5\textwidth]{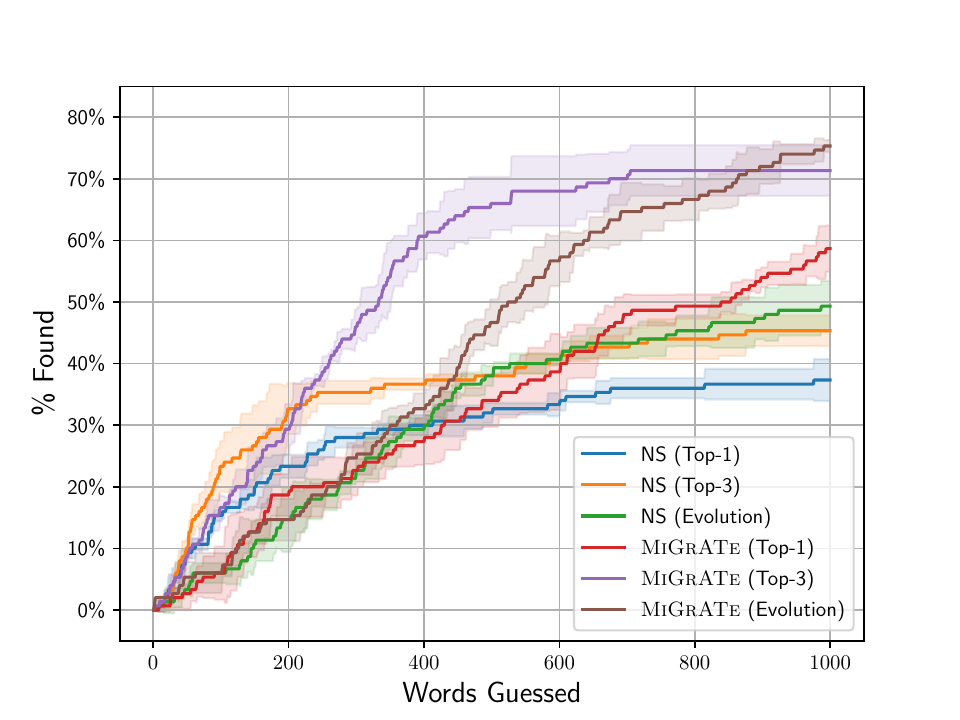}
    \caption{\textbf{Comparing selection methods for NS.} Evolution-based selection shows slower initial gains but results in more consistent improvements than using a top-$k$ sampling strategy--resulting in better final performances.}
    \label{fig:semantle_evolution}
\end{figure}

We conduct a comparison of using NS and \method{} with three different strategies for selecting the solution to sample neighbors from: Top-1, Top-3, and Evolution. For each of these configurations we use 10 neighborhood samples, 0 online samples, and 0 greedy samples. Fig.~\ref{fig:semantle_evolution} shows that Top-3 outperforms Top-1 and that using our evolution-based strategy outperforms Top-3 in both NS and \method{} methods. While Top-3 shows the better initial gains in both NS and \method{}, the evolution-based strategy narrowly outperforms it by 1000 samples.  Much like our other results in Table.~\ref{tab:combined-table}, we also observe that the \method{} equivalent of each NS variation performs better -- reinforcing the pattern that TTT improves search performance. 

\subsection{Can related tasks bootstrap search?}
We investigate whether fine-tuned weights from TTT can generalize to other tasks. After running \method{} on every task, we perform TTT again on unsolved tasks and bootstrap the method with the learned weights of its ``nearest'' solved task.

In this experiment, we attempt to solve ARC tasks that were not solved by \method{}. For each unsolved task, we determine its ``nearest'' solved task by evaluating this task using the solution program from every solved task. We pass the training inputs of the unsolved task into each program and determine the nearest solved task to be the one whose solution program achieve the highest reward from our hamming distance-based reward function.

Once the nearest solved task is identified, we use its fine-tuned weights from \method{} as the initializing point for solving the unsolved task. This procedure aims to transfer inductive biases that may have been learned from structurally similar tasks, enabling the model to efficiently explore more viable programs on the unsolved task. This tests whether there is an advantage to initializing search via TTT from a more informed starting point on problems where starting with the base model fails. 

\begin{figure}[htbp]
    \centering
    \includegraphics[width=0.5\textwidth]{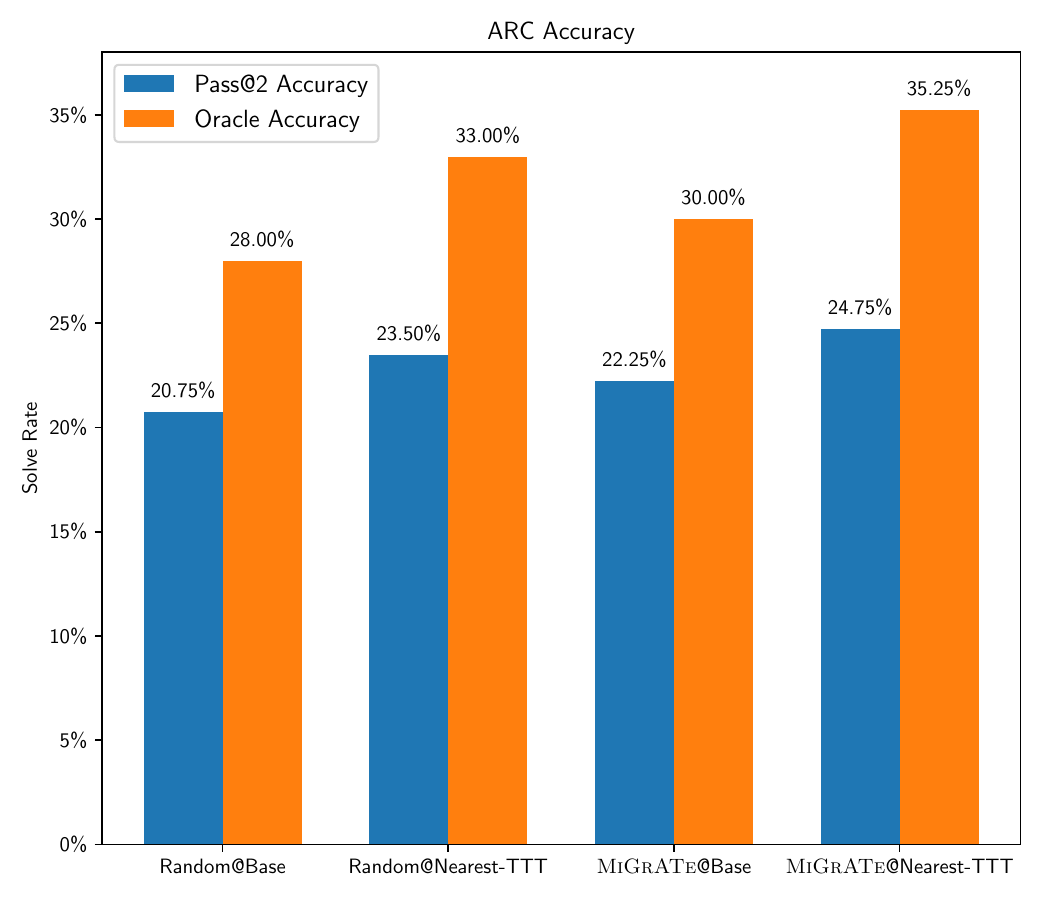}
    \caption{\textbf{Bootstrapping with nearest weights on ARC-Full.} Bootstrapping Random and \method{} with initial weights learned from one round of \method{} shows slight improvement on total tasks solved.}
    \label{fig:bootstrap}
\end{figure}

We see marginal improvements from bootstrapping search with learned weights from \method{}. Fig. ~\ref{fig:bootstrap} shows that initializing Random Sampling and \method{} with the nearest solved task's weights allowed each respective method to solve tasks that were initially unsolvable by the base model. Notably, bootstrapping Random Sampling with nearest weights was able to solve more tasks than executing \method{} on the base model.

\begin{figure*}[t]
    \centering
    \subfloat[Semantle]{%
        \includegraphics[width=0.32\textwidth]{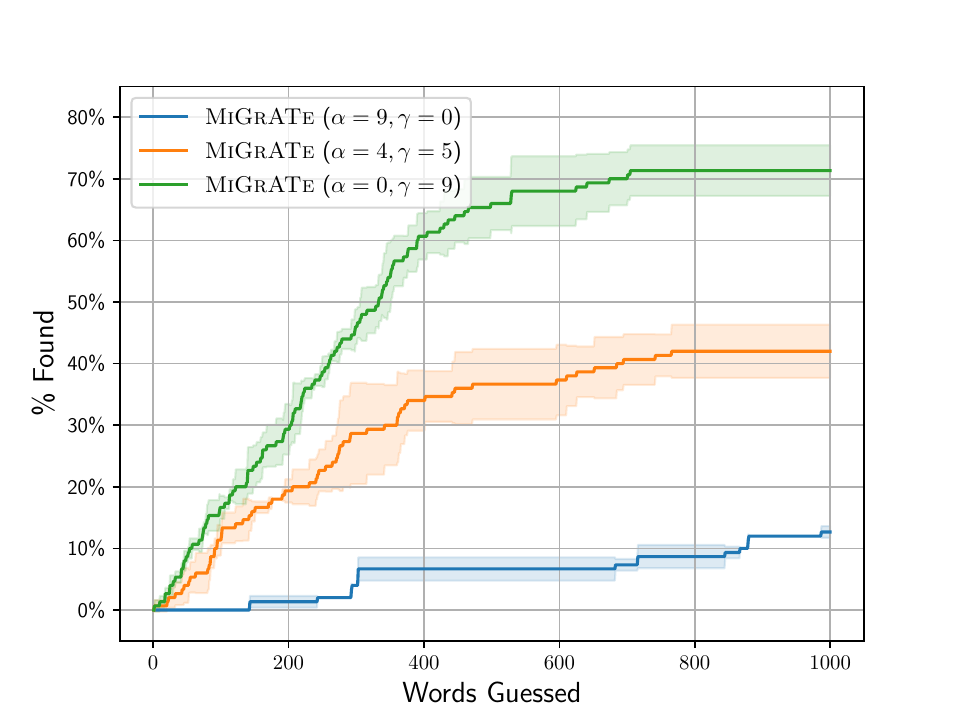}
    }
    \hfill
    \subfloat[Dockstring]{%
        \includegraphics[width=0.32\textwidth]{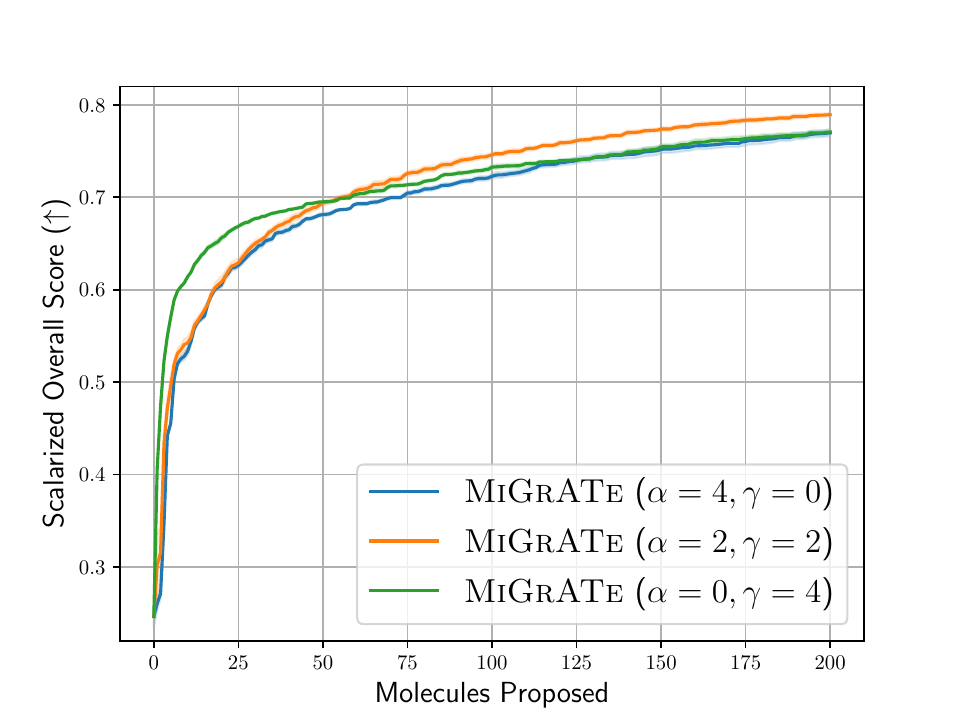}
    }
    \hfill
    \subfloat[ARC-Small]{%
        \includegraphics[width=0.32\textwidth]{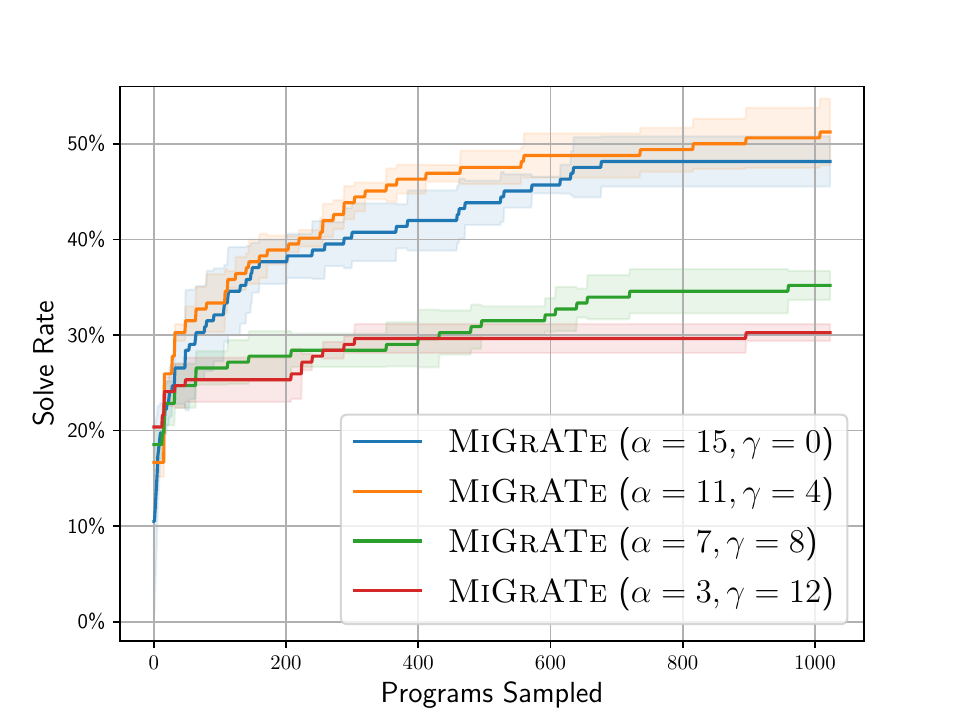}
    }
    \caption{\textbf{Varying \nonline{} and \nnbr{}.} We vary the number of online and NS samples per group in \method{}. \textbf{(a)} On Semantle, we found that the strategy of using no online samples to be the most successful by a significant margin. \textbf{(b)} On Dockstring, we found that using only NS samples yield better performances at smaller budgets and a configuration of equal amounts of online and NS samples to achieve the best final performance. \textbf{(c)} On ARC-Small, we found the mixed configuration of $\alpha=11$ and $\gamma=4$ to perform the best.}
    \label{fig:vary_ns_plots}
\end{figure*}

\subsection{Tradeoff with Varying \nonline{} and \nnbr{} Samples}
We conduct experiments on Semantle, Dockstring, and ARC-Small to investigate the tradeoff involved in varying the ratio of online to neighborhood samples within a GRPO group in \method{}. Throughout these experiments, we fix the number of greedy samples at $\beta=1$. The results in Fig.~\ref{fig:vary_ns_plots} reveals that the optimal configuration of online sand NS samples vary across domains. Particularly, Semantle benefits from more NS samples, Dockstring performs the best with an equal ratio of samples, while ARC prefers a higher proportion of online samples. These results highlights the importanced of tuning \nonline{} and \nnbr{} when applying \method{} to different domains.

\subsection{Varying \ngreedy{} Samples}

\begin{figure}[htbp]
    \centering
    \includegraphics[width=0.5\textwidth]{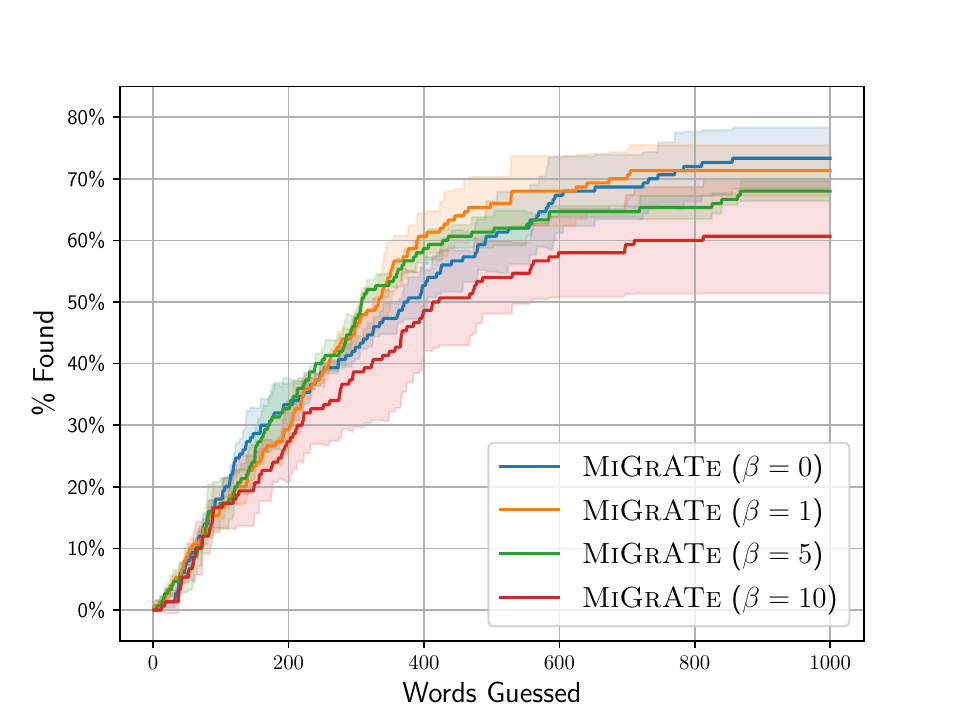}
    \caption{\textbf{Comparing $\beta$ on Semantle.} \method{} shows a bias towards smaller \ngreedy{} for better performance on Semantle.}
    \label{fig:semantle_vary_beta}
\end{figure}

We explore varying the number of greedy samples on Semantle. In these experiments, we run \method{} with $\alpha = 0$ onlines amples, $\beta$ greedy samples, and $\groupsize - \beta$ neighborhood sampless. As shown in Fig.~\ref{fig:semantle_vary_beta},  performance remains relatively similar over $\ngreedy{}=0,1,5,10$ with a small trend of better performance with smaller $\ngreedy{}$. In tandem with the results on varying \nnbr{}, this supports the potential of more off-policy methods of performing TTT with GRPO.

\begin{table*}[htbp]
  \centering
  \hspace*{-5.5em}
  \footnotesize
  \begin{tabular}{l c ccc cc} \toprule
    & \multicolumn{1}{c}{\textbf{Semantle}} & \multicolumn{3}{c}{\textbf{Dockstring}}  & \multicolumn{2}{c}{\textbf{ARC-Small}}\\
    \cmidrule(lr){2-2}
    \cmidrule(lr){3-5}
    \cmidrule(lr){6-7}
    \textbf{Method} & \textbf{\% Found} & \textbf{QED ($\uparrow$)} & \textbf{Vina Score ($\downarrow$)} & \textbf{Overall Score ($\uparrow$)} & \textbf{Pass@2 (\%)} & \textbf{Oracle (\%)}\\
    \midrule
    NS & \secondbestresult{45.30 \pm 2.49} & $0.87 \pm 0.01$ & $-9.65 \pm 0.21$ & $0.71 \pm 0.00$ & $48.15 \pm 0.00$ & $55.56 \pm 1.51$\\
    OPRO & $40.70 \pm 1.89$ & $0.90 \pm 0.00$ & $-9.94 \pm 0.06$ & $0.74 \pm 0.00$ & \secondbestresult{50.62 \pm 1.75} & \secondbestresult{59.26 \pm 0.00}\\
    \midrule
    \method{} & \bestresult{71.30 \pm 4.11} & \bestresult{0.90 \pm 0.00} & \bestresult{-11.00 \pm 0.07} & \bestresult{0.79 \pm 0.00} & \bestresult{51.23 \pm 3.49} & \bestresult{62.35 \pm 0.87}\\
    \method{} (OPRO) & \secondbestresult{65.3\% \pm 2.49} & \bestresult{0.90 \pm 0.00} & \secondbestresult{-10.80 \pm 0.10} & \secondbestresult{0.78 \pm 0.00} & $44.44\% \pm 3.02$ & $55.56 \pm 0.04$ \\
    
    \bottomrule
  \end{tabular}
\caption{\textbf{Comparing Prompt Optimization Techniques.} We compare the inference-only and \method{} (TTT) performance of different prompt optimization techniques. All results are averaged over three random seeds, with the standard deviation reported. The best result in each column is marked in bold and the second best result is underlined. \method{} achieves the best performance across all metrics and ties with \method{} (OPRO) on optimizing QED for Dockstring. Notably, OPRO beats NS in every metric with the exception of accuracy on Semantle.}
\label{tab:opro-table}
\end{table*}

\begin{figure}[htbp]
    \centering
    \subfloat[Semantle]{%
        \includegraphics[width=0.32\textwidth]{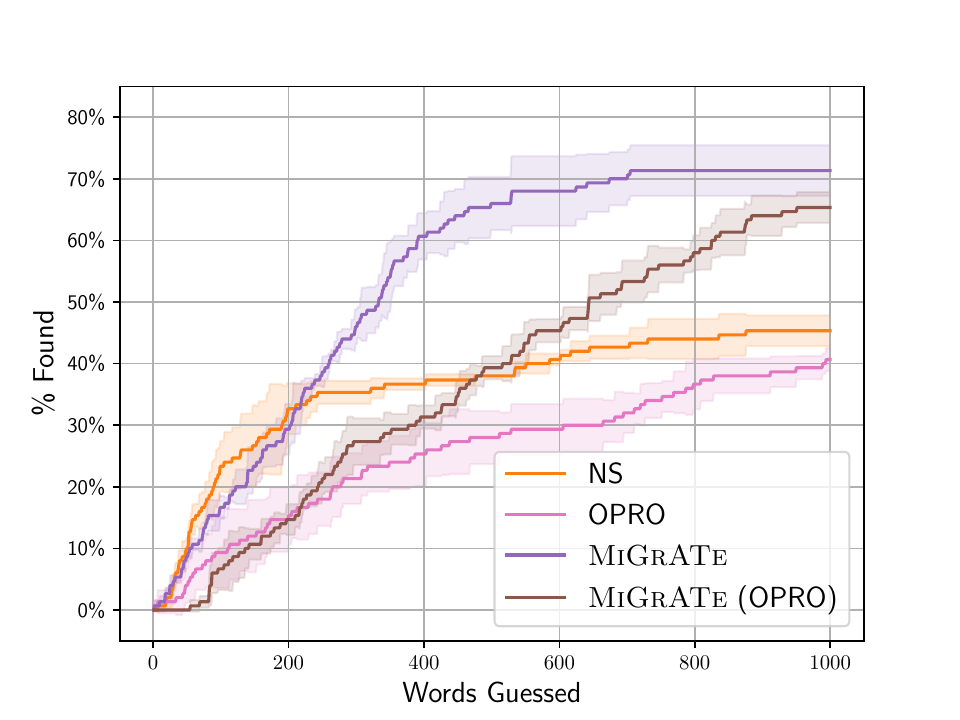}
    }
    \hfill
    \subfloat[Dockstring]{%
        \includegraphics[width=0.32\textwidth]{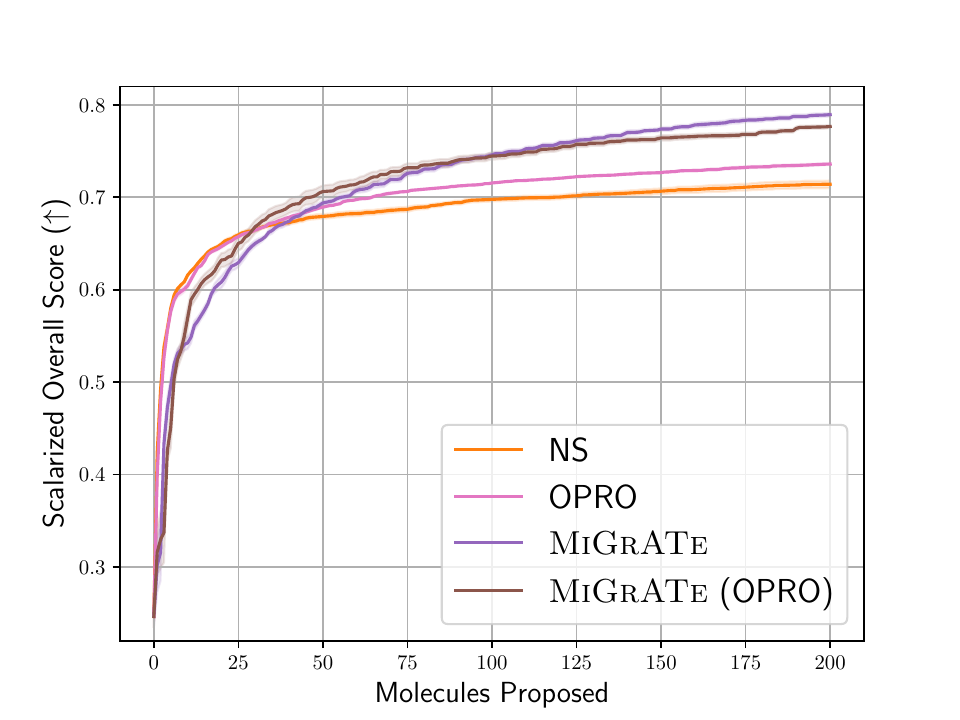}
    }
    \hfill
    \subfloat[ARC-Small]{%
        \includegraphics[width=0.32\textwidth]{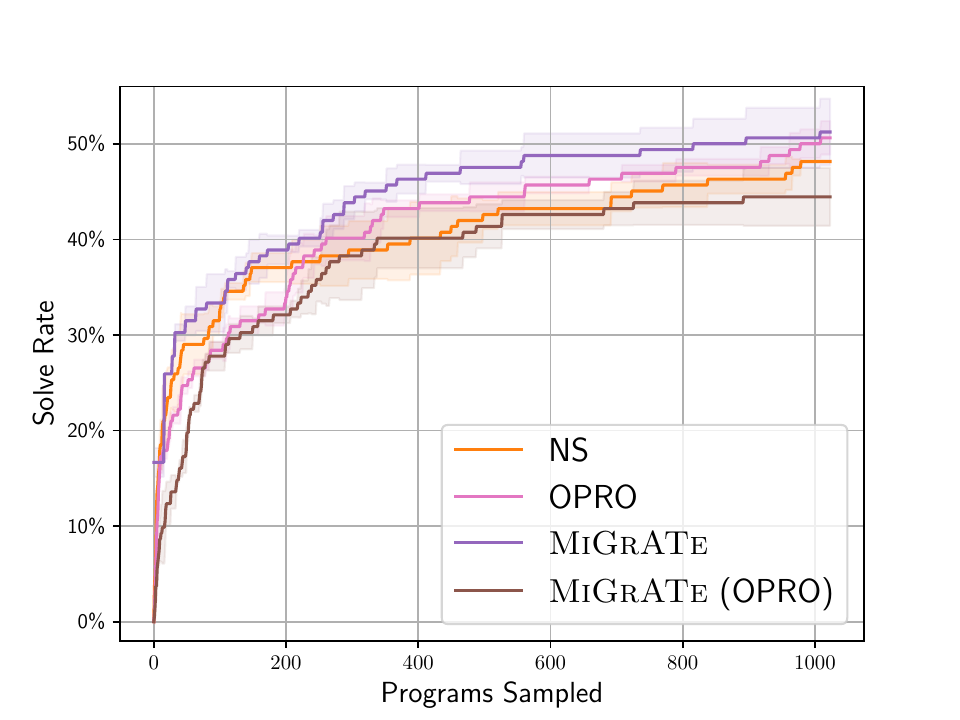}
    }
    \caption{\textbf{Comparing Prompt Optimization Techniques.} \method{} (OPRO) shows similar performance to \method{} on Semantle and Dockstring and noticeably worse performance on ARC-Small.}
    \label{fig:migrate_opro}
\end{figure}

\begin{figure}[htbp]
    \centering
    \subfloat[Best-so-far QED]{%
        \includegraphics[width=0.49\textwidth]{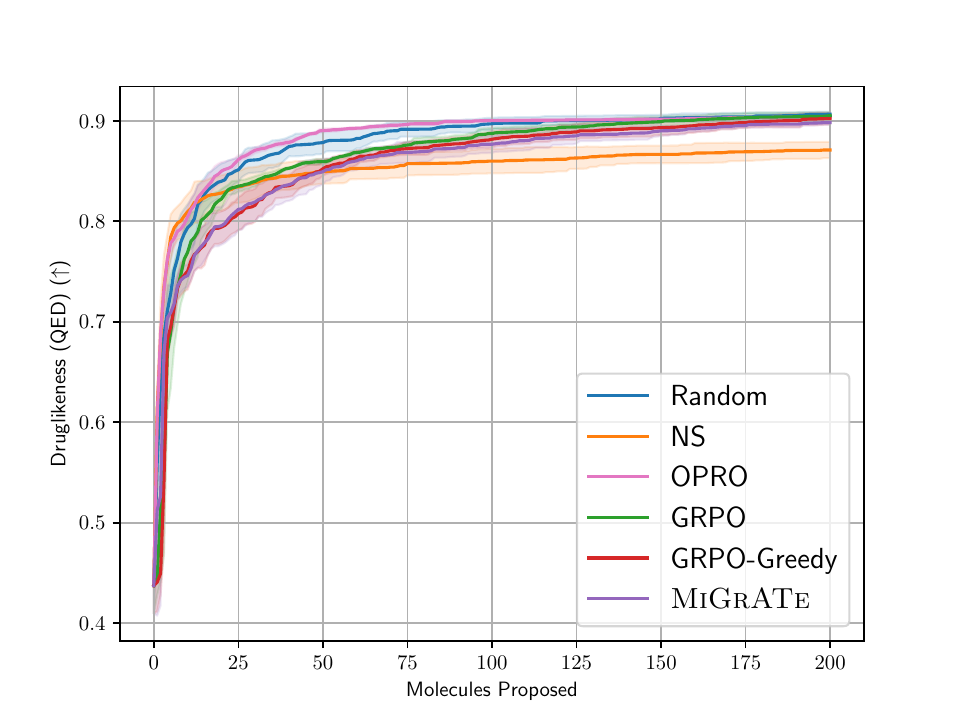}
    }
    \hfill
    \subfloat[Bset-so-far Vina]{%
        \includegraphics[width=0.49\textwidth]{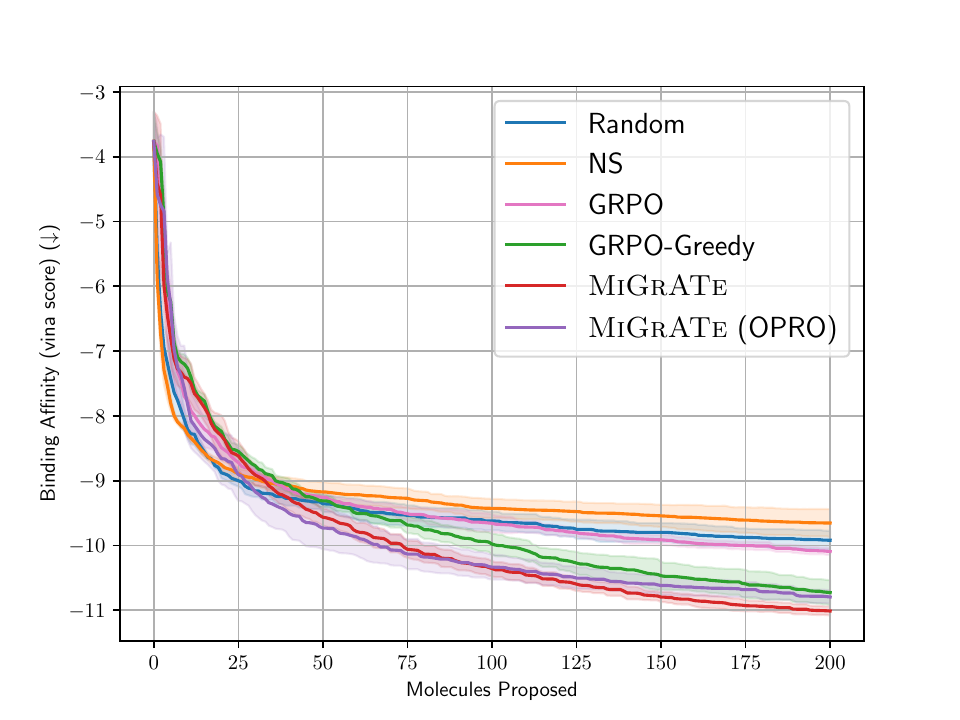}
    }
    \caption{\textbf{QED and Vina Score plots for Dockstring.}}
    \label{fig:mol-qed-vina}
\end{figure}

\subsection{Alternative local structure sampling in \method{}?}
\label{appendix:migrate_opro}

We experiment with the alternative of using OPRO in place of neighborhood sampling (NS) in \method{}. Our results in Table.~\ref{tab:opro-table} and Fig.~\ref{fig:migrate_opro} show similar results between \method{} and \method{} (OPRO) on Dockstring and more favorable results towards \method{} on Semantle and ARC-Small. Compared to other baselines in Table~\ref{tab:combined-table}, \method{} (OPRO) only underperforms relative to \method{} on Semantle and Dockstring. Notably, on ARC-Small, incorporating TTT into OPRO substantially degrades performance compared to inference-only OPRO. We also observe that OPRO achieves better performance than NS across most metrics. The varying performance of \method{} (OPRO) across domains suggests that NS is more compatible than OPRO with \method{}. In addition, the greater improvement achieved by using NS over OPRO suggests that the NS strategy of generating diverse variations may be better suited to TTT than OPRO, which focuses more on direct improvement of previous solutions.

\onecolumn

\section{Appendix C: LLM Prompts}
\label{appendix:appendix_c}

\label{appendix:llm_prompts}

\subsection{Semantle: Task Prompt}
\label{appendix:task_prompts}

\begin{tcolorbox}[colback=yellow!10, colframe=black, arc=2mm, boxrule=0.5pt, width=0.95\linewidth, left=0.5pt, right=0.5pt, breakable, enhanced]
\ttfamily\frenchspacing
    Your task is to guess a hidden word from the English dictionary. Stick to proper, single-word English words. Now, guess exactly n=\%s new word(s) that could be the hidden word. Be creative! (Note: give only a list of word(s) in the provided JSON format, e.g. {"response": ["word1", "word2",...]})
\end{tcolorbox}

\vspace{0.5em}

\subsection{Semantle: Neighborhood Sampling Prompt}

\begin{tcolorbox}[colback=yellow!10, colframe=black, arc=2mm, boxrule=0.5pt, width=0.95\linewidth, left=0.5pt, right=0.5pt, breakable, enhanced]
\ttfamily\frenchspacing
    Your task is to guess words related to a word from the English dictionary. Stick to proper, single-word English words. Now, guess exactly n=\%s new word(s) that could be related to the word(s):
    \medskip

    Word: \%s
    \medskip
    
    Be creative! (Note: give only a list of word(s) in the provided JSON format, e.g. {"response": ["word1", "word2",...]})
\end{tcolorbox}

\vspace{0.5em}

\subsection{Dockstring: Task Prompt}

\begin{tcolorbox}[colback=yellow!10, colframe=black, arc=2mm, boxrule=0.5pt, width=0.95\linewidth, left=0.5pt, right=0.5pt, breakable, enhanced]
\ttfamily\frenchspacing
 Your task is to find the optimal drug molecule that has both a high druglikeness (QED) as well as a strong binding affinity (vina) with the protein \%s. For docking, lower is better (less than --10 is considered good) and for druglikeness, 1 is the best and 0 is the worst (greater than 0.8 is considered good). While both properties are important, the docking score is 10 times as important as the druglikeness score. If you propose an invalid molecule or make a repeat guess, you will get no score, so stick to valid SMILES strings.

\medskip

Now, guess exactly n=\%s new molecule(s).

\medskip

(Note: give only a list of SMILES string(s) in the provided JSON format, e.g. {"response": ["SMILES1", "SMILES2", ...]})
\end{tcolorbox}

\vspace{0.5em}

\subsection{Dockstring: Neighborhood Sampling Prompt}

\begin{tcolorbox}[colback=yellow!10, colframe=black, arc=2mm, boxrule=0.5pt, width=0.95\linewidth, left=0.5pt, right=0.5pt, breakable, enhanced]
\ttfamily\frenchspacing
    Your task is to find the optimal drug molecule that has both a high druglikeness (QED) as well as a strong binding affinity (vina) with the protein \%s. For docking, lower is better (less than --10 is considered good) and for druglikeness, 1 is the best and 0 is the worst (greater than 0.8 is considered good). While both properties are important, the docking score is 10 times as important as the druglikeness score. If you propose an invalid molecule or make a repeat guess, you will get no score, so stick to valid SMILES strings!
    
\medskip

Here is my guess for a molecule:

SMILES: \%s

\medskip

Now, guess exactly n=\%s new variation(s) of my molecule that could improve the scores to reach the optimal molecule.

\medskip

(Note: give only a list of SMILES string(s) in the provided JSON format, e.g. {"response": ["SMILES1", "SMILES2", ...]})
\end{tcolorbox}

\subsection{ARC: Task Prompt}

\begin{tcolorbox}[colback=yellow!10, colframe=black, arc=2mm, boxrule=0.5pt, width=0.95\linewidth, left=0.5pt, right=0.5pt, breakable, enhanced]
\ttfamily\frenchspacing
    Given input-output grid pairs as reference examples, carefully observe the patterns to predict the output grid for new test input. Each pair follows the same transformation rule. Grids are 2D arrays represented as strings, with cells (colors) separated by spaces and rows by newlines. Here are the input and output grids for the reference examples:
    
\medskip

Example 1:

Input:

[[1,1,1,...,1]]

Output:

[[2,2,2,...,2]]

\medskip

Example 2:

Input:

[[2,2,2,...,2]]

Output:

[[3,3,3,...,3]]

\medskip

...

\medskip
Here is the input grid for the test example:

Input:

[[3,3,3,...,3]]
\medskip

Write a Python function `transform` that can convert any given input grid to its corresponding output grid based on the pattern observed in the reference examples.
\end{tcolorbox}

\vspace{0.5em}

\subsection{ARC: Neighborhood Sampling Prompt}

\begin{tcolorbox}[colback=yellow!10, colframe=black, arc=2mm, boxrule=0.5pt, width=0.95\linewidth, left=0.5pt, right=0.5pt, breakable, enhanced]
\ttfamily\frenchspacing
Given input-output grid pairs as reference examples, carefully observe the patterns to predict the output grid for new test input. Each pair follows the same transformation rule. Grids are 2D arrays represented as strings, with cells (colors) separated by spaces and rows by newlines. 

\medskip

Here are the input and output grids for the reference examples:
    
\medskip

Example 1:

Input:

[[1,1,1,...,1]]

Output:

[[2,2,2,...,2]]

\medskip

...

\medskip
Here is the input grid for the test example:

Input:

[[3,3,3,...,3]]

\medskip

The goal is to write a Python function `transform` that can convert any given input grid to its corresponding output \
grid based on the pattern observed in the reference examples.

\medskip

Here is my guess for the function:

```python

def transform(input: np.ndarray) -> np.ndarray:

\hspace*{2em}\# Code

```

\medskip

Provide a variation of my guess that could be the correct answer.
\end{tcolorbox}

\end{document}